\PassOptionsToPackage{table}{xcolor}
\documentclass[10pt,twocolumn,letterpaper]{article}

\usepackage{iccv}            
\definecolor{iccvblue}{rgb}{0.21,0.49,0.74}
\usepackage[pagebackref,breaklinks,colorlinks,allcolors=iccvblue]{hyperref}
\usepackage{placeins} 
\usepackage{float}
\usepackage{multirow}
\usepackage{stfloats}
\usepackage{xcolor} 
\usepackage{caption}
\usepackage{capt-of}

\title{EquiCaps: Predictor-Free Pose-Aware Pre-Trained Capsule Networks}

\author{Athinoulla Konstantinou$^{1,2}$\thanks{Corresponding author. Email: \href{mailto:a.konstantinou.24@abdn.ac.uk}{a.konstantinou.24@abdn.ac.uk}.}  \and Georgios Leontidis$^{1,2}$  \and Mamatha Thota$^{3}$  \and Aiden Durrant$^{1}$   \and \normalfont  
\small$^1$School of Natural and Computing Sciences, \\ \small University of Aberdeen, UK    \and 
\normalfont \and 
\hspace{-3em} \small$^2$Interdisciplinary Institute, \\ \small \hspace{-3.7em}  University of Aberdeen, UK \and
\small  $^3$School of Computer Science, \\ \small \hspace{0.4em}  University of Lincoln, UK 
}

\begin{document}

\maketitle
\begin{abstract}
Learning self-supervised representations that are invariant and equivariant to transformations is crucial for advancing beyond traditional visual classification tasks. However, many methods rely on predictor architectures to encode equivariance, despite evidence that architectural choices, such as capsule networks, inherently excel at learning interpretable pose-aware representations. To explore this, we introduce EquiCaps (Equivariant Capsule Network), a capsule-based approach to pose-aware self-supervision that eliminates the need for a specialised predictor for enforcing equivariance. Instead, we leverage the intrinsic pose-awareness capabilities of capsules to improve performance in pose estimation tasks. To further challenge our assumptions, we increase task complexity via multi-geometric transformations to enable a more thorough evaluation of invariance and equivariance by introducing 3DIEBench-T, an extension of a 3D object-rendering benchmark dataset. 
Empirical results demonstrate that EquiCaps outperforms prior state-of-the-art equivariant methods on rotation prediction, achieving a supervised-level $R^2$ of 0.78 on the 3DIEBench rotation prediction benchmark and improving upon SIE and CapsIE by 0.05 and 0.04 $R^2$, respectively. 
Moreover, in contrast to non-capsule-based equivariant approaches, EquiCaps maintains robust equivariant performance under combined geometric transformations, underscoring its generalisation capabilities and the promise of predictor-free capsule architectures.
Code, dataset, and weights are released at \url{https://github.com/AberdeenML/EquiCaps}.
\end{abstract}

\section{Introduction}
\label{sec:intro}

The success of self-supervised learning (SSL) is largely attributed to the exploitation of \textit{invariant} representations under data augmentations and transformations to learn rich semantic concepts~\cite{chen2020simple,he2020momentum, chen2020improved, grill2020bootstrap, caron2020unsupervised, zbontar2021barlow, bardes2022vicreg, caron2021emerging, he2022masked, bardes2022vicregl}. However, methods that strictly enforce invariance fail to preserve useful augmentation and transformation information, which could limit their applicability to downstream tasks where such information is vital~\cite{whatshouldnotbecontrastive, lee2021improving, gupta2024structuring}. In response, there has been increasing interest in developing more generalised self-supervised approaches, most notably through the investigation of \textit{pose-aware} learning~\cite{SIE, wang2024pose, wang2025understanding}. Achieving such generalisation, particularly to novel viewpoints, can consequently address real-world settings that inherently exhibit large data variations, \eg 
autonomous driving~\cite{wang2024pose}. 

\begin{figure}[t!]
  \centering
    \includegraphics[width=0.9\linewidth]{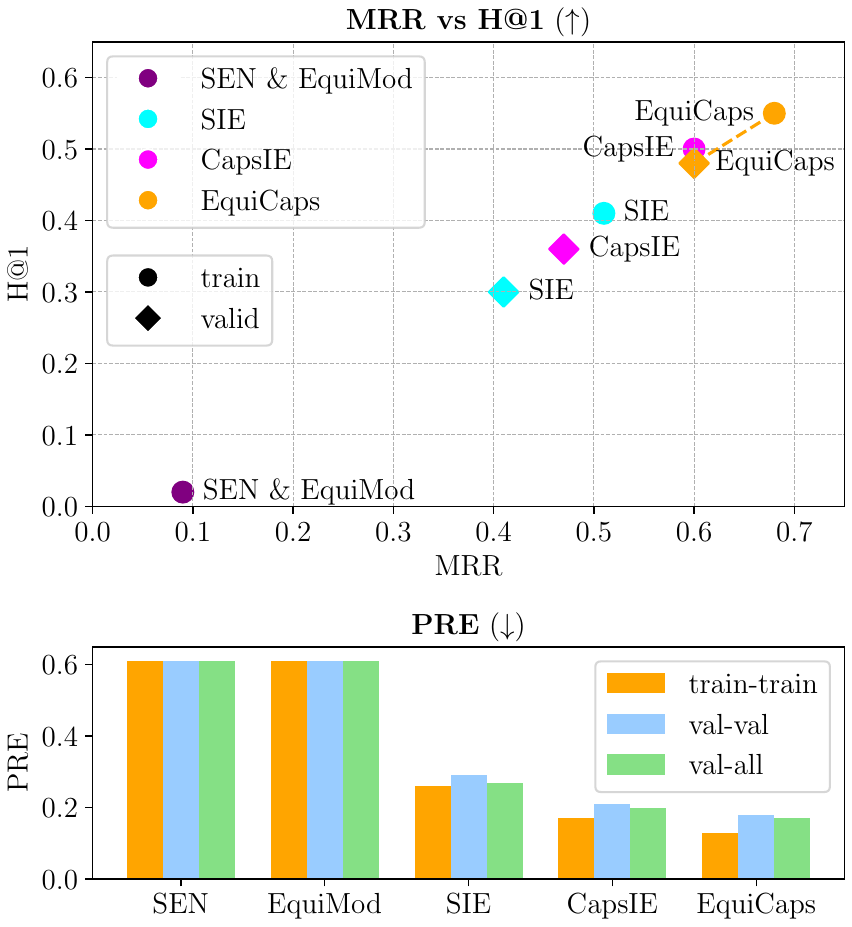}
    \caption{Equivariant evaluation of EquiCaps (ours) and competing methods on 3DIEBench benchmark.
     We indicate which subset (train or val) is used to generate embeddings, and, for PRE, which dataset (train, val, or all) is used for retrieval.
    }
    \label{fig:Fig-1}
\end{figure}

One promising strategy to preserve transformation information—and thus achieve pose-awareness—is to directly optimise for \textit{equivariant} representations, enabling the model to capture how transformations steer the latent space. However, such methods often face limitations~\cite{park2022sen, equimod, SIE, capsie}. First, they rely on a dedicated predictor network to encode equivariant properties, often further increasing model complexity through conditional projectors, partitioned representations, or hypernetworks. Second, equivariance in representations is typically enforced solely via objective functions with little regard taken to exploit architectural mechanisms—such as Capsule Networks (CapsNets)—that have been shown to effectively encode or structure pose information~\cite{capsules_survey}.

We propose to simplify and enhance the framework of equivariant self-supervision to address these concerns by using CapsNets~\cite{capsules_survey}. CapsNets structure representations into capsules, each encoding both the existence of an entity and its instantiation parameters, where the latter is achieved through a routing algorithm based on agreement of part-whole relationships. This formulation introduces an inductive bias that has been shown to effectively capture equivariance w.r.t. viewpoints and invariance in the capsule activations, enabling them to exhibit viewpoint-invariant and viewpoint-equivariant properties by design. 

While prior attempts have been made to introduce CapsNets into equivariant SSL~\cite{capsie}, they still rely on a separate transformation predictor, potentially underusing capsules' intrinsic pose-awareness capabilities. To overcome these limitations, we propose a streamlined pipeline that eliminates the predictor and directly leverages capsules' equivariant properties that prior equivariant SSL methods do not. CapsNets offer an intriguing prospect for capturing transformations, therefore we introduce an extension of the 3DIEBench dataset~\cite{SIE} that incorporates translational transformations in addition to rotations. By moving from rotation-only SO(3) transformations~\cite{SIE, wang2024pose} to the more general SE(3) group, our dataset increases the difficulty of both invariance and equivariance tasks, allowing for the evaluation of generalisation under multi-geometric settings.
To summarise, our contributions are as follows: 
\begin{itemize}[noitemsep]
\item \textbf{3DIEBench-T:} We present 3DIEBench-T, an extension of the 3DIEBench that integrates translational transformations alongside rotations. This new benchmark enables the evaluation of representation methods under simultaneous multi-geometric 3D transformations.
\item \textbf{Predictor Limitations:}
We empirically demonstrate that existing predictor-based equivariant architectures yield minimal improvement over purely invariant approaches when learning multi-geometric 3D transformations.
\item \textbf{Predictor-Free CapsNet:} We propose EquiCaps, removing the need for predictors by exploiting the intrinsic pose-awareness of CapsNets. Empirical results on rotation and translation tasks demonstrate that EquiCaps outperforms existing equivariant methods, and generalises successfully under multi-geometric settings, while also allowing intuitive manipulation of the embedding space.

\end{itemize}

\section{Preliminaries}
\label{sec:preliminaries}

\textbf{Group actions.}
A group is a set \( G \) with a binary operation \( \cdot \) : \(  G \times G \to G \) that satisfies associativity, the existence of an identity element $e$, \( g \cdot e = e \cdot g = g \  \), \( \forall 
g \in G \), and the existence of an inverse $g^{-1}$, \( g \cdot g^{-1} = g^{-1} \cdot g = e \), \( \forall 
g \in G \).
A group action of \( G \) on a set \( S \) is defined as a map \( \alpha : G \times S \to S \) such that $\alpha(e,s) = s$ and that is compatible with the composition of group elements, \( \alpha(g, \alpha(h, s)) = \alpha(gh, s) \), \( \forall s \in S \) and \( \forall g, h \in G \). 
$\alpha$ is called a \textit{group representation} when acting on a vector space \( V \), such as \( \mathbb{R}^n \), and \( \alpha \) is linear. A group representation is denoted as the map \( \rho : G \to GL(V) \), such that \( \rho(g) = \alpha(g, \cdot) \). 
As in similar works~\cite{SIE}, we use group representations to describe how transformations are applied both to the input data and latent space.

\textbf{Equivariance Learning.}
A function \( f : X \to Y \) is equivariant if for a group \( G \), with representations \( \rho_X \) and \( \rho_Y \) on \( X \) and \( Y \), satisfies  
 \begin{equation}
f(\rho_X(g) \cdot x) = \rho_Y(g) \cdot f(x),   \forall x \in X, \forall g \in G. 
\label{eq:equivariance}
\end{equation}

\noindent This indicates the commutativity of group transformations and function application. 

\textbf{Invariance Learning.}
 \noindent A function \( f : X \to Y \) is invariant if for a group \( G \), with representations \( \rho_X \) on \( X \), satisfies 
\begin{equation}
f(\rho_X(g) \cdot x) = f(x), \quad \forall x \in X, \forall g \in G.
\label{eq:invariance}
\end{equation}
Invariance is a trivial, special case of equivariance when \( \rho_Y(g)\), in Eq.~\eqref{eq:equivariance}  is the identity map.
\section{Related  Works} 
\label{sec:relatedwork}

\textbf{Invariant SSL.} 
Invariance to data augmentations has been the driving force of self-supervision, using this property to learn common semantic concepts under different transformations when there is no access to labels. Specifically, invariance in the latent space is enforced by aligning shared views that are transformations of the same source. There are various approaches to undertake this alignment and discrimination, notably contrastive methods~\cite{chen2020simple, he2020momentum,chen2020improved,chen2021empirical,yeh2022decoupled}, clustering-based learning~\cite{caron2020unsupervised, caron2021emerging, caron2018deep, oquab2024dinov}, and non-contrastive methods~\cite{grill2020bootstrap,chen2021exploring, bardes2022vicreg, zbontar2021barlow, ermolov2021whitening, bardes2022vicregl} which instead employ architectural or objective modifications, gradient mechanisms, and information-theoretic regularisers.

\textbf{Implicit Equivariant SSL.}
Invariant SSL methods are optimised to disregard data transformations, hence information is lost. To address this, predictive equivariant SSL encodes transformation-sensitive information by training to predict transformation parameters~\cite{scherr2022selfsupervised, lee2021improving, gidaris2018unsupervised}, while others integrate contrastive learning with transformation prediction~\cite{dangovski2021equivariant}. Furthermore, modulating the contrastive objective itself, based on the strength of the applied augmentation has shown promise~\cite{Xie_2022_CVPR}, while~\cite{whatshouldnotbecontrastive} introduces distinct embedding spaces that remain invariant to all but one augmentation. Focussing on the transformation, optimal augmented views can also be exploited to retain task-relevant information~\cite{NEURIPS2020_4c2e5eaa}. However, these are not truly equivariant since they do not learn an explicit mapping from input transformations to corresponding changes in latent space~\cite{wang2025understanding}.

\textbf{Explicit Equivariant SSL.}
Avoiding transformation prediction tasks in favour of evaluating for equivariant representations, elicits a separate family of methods. These utilise representation reconstruction and maintain invariant and equivariant properties by splitting into or learning separate latent spaces~\cite{winter2022unsupervised, marchetti23b}. Others~\cite{Shakerinava, stl} encourage equivariance by sampling paired images under the same transformations, with~\cite{Shakerinava} utilising the transformation group information for choosing optimal loss terms. Such approaches have also extended contrastive learning enforcing input transformations correspond to rotations in the embedding space~\cite{gupta2024structuring}. Several methods adopt a predictor network to enforce an equivalent transformation mapping between latent representations through augmentation parameters~\cite{equimod, park2022sen}. Building on these, SIE~\cite{SIE} incorporates a hypernetwork to learn such mapping, while dividing the representation into invariant and equivariant parts. CapsIE~\cite{capsie} adopts this module, instead leveraging a CapsNet projector, removing the need for multiple projectors and the split representation. 

\textbf{CapsNets.} CapsNets are designed to model part–whole hierarchical relationships by capturing the likelihood of an entity's presence and its attributes (pose, colour, texture, etc.) through neural activity. Lower-level capsules connect to higher-level ones via routing based on agreement in likelihood and positional attributes. This alignment enables capsules to identify objects by reinforcing pose consistency when multiple capsules converge on the same entity. Due to this positional agreement mechanism, CapsNets are explored as equivariant architectures, enabling tasks like image generation~\cite{hintontransformingautoencoders} and transformation learning~\cite{sabour2017dynamic}. Various routing methods have been proposed, including Gaussian-mixture clustering~\cite{hinton2018matrix}, Variational Bayes with uncertainty~\cite{ribeiro2020capsule}, self-routing via subordinate networks~\cite{hahn2019self}, and residual pose routing~\cite{liu2024capsule}.

\section{3DIEBench-T: A Translation-Enhanced Benchmark for Invariant-Equivariant SSL}
\label{sec:3diebench-t}

As highlighted by~\citet{SIE}, commonly used datasets for evaluating invariant or equivariant properties face limitations. Evaluations of invariant representations commonly benchmark on datasets like ImageNet~\cite{deng2009imagenet} or CIFAR~\cite{krizhevsky2009learning}, where objects lack parameterisable transformations because augmentations occur only in the pixel domain. However, certain equivariant datasets~\cite{Kipf2020Contrastive, park2022sen} lack visual complexity and transformational diversity to allow for fine-grained control.
A more comprehensive alternative is 3DIEBench~\cite{SIE}, which controls the scene and object parameters for 3D rotation prediction (equivariance) while retaining sufficient complexity for classification (invariance). Nonetheless, as with other similar datasets~\cite{wang2024pose}, 3DIEBench focuses only on rotation, evaluating methods under a single geometric transformation.

Focussing only on rotation raises questions regarding the generalisability of methods~\cite{SIE, capsie} to other geometric transformations, and their applicability in more realistic settings. To address this, we extend 3DIEBench by introducing additional translational transformations investigating transformations applied to both base and object frames of reference. Our new dataset, 3DIEBench-T (3D Invariant Equivariant Benchmark–Translated), goes beyond rotation-only benchmarks~\cite{SIE,wang2024pose}, enabling translation prediction and providing a more challenging, realistic setting to assess how methods handle simultaneous multi-geometric transformations.

To investigate the impact of the added transformations, we adhere to the original 3DIEBench data generation protocol, ensuring any observed performance differences are attributed to the inclusion of translations rather than broader dataset modifications.
We use 52,472 3D object instances spanning 55 classes from ShapeNetCoreV2~\cite{shapenet}, originally sourced from 3D Warehouse~\cite{trimble3dwarehouse}.
For each instance, we generate 50 uniformly sampled views within specified ranges via Blender-Proc~\cite{denninger2019blenderproc}, yielding 2,623,600 images.
Samples appear in \cref{fig:3DIEBench-T}, with further generation information and visualisations given in the supplementary Sec.~\ref{sec:dataset-generation}. 

\begin{figure}[h!]
  \centering
    \includegraphics[width=1\linewidth]{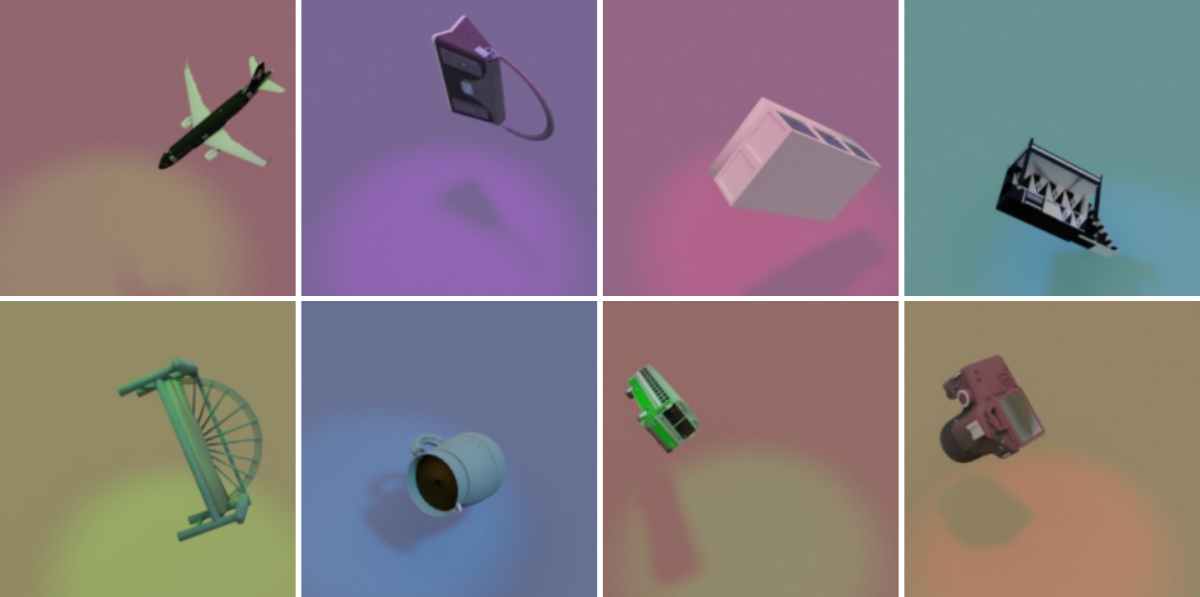}
    \caption{Sample object instances from 3DIEBench-T dataset under different rotation, translation, and colour hue transformations.}
    \label{fig:3DIEBench-T}
\end{figure}

\begin{figure*}[t]
    \centering
    \includegraphics[width=\linewidth,height=7cm,keepaspectratio]{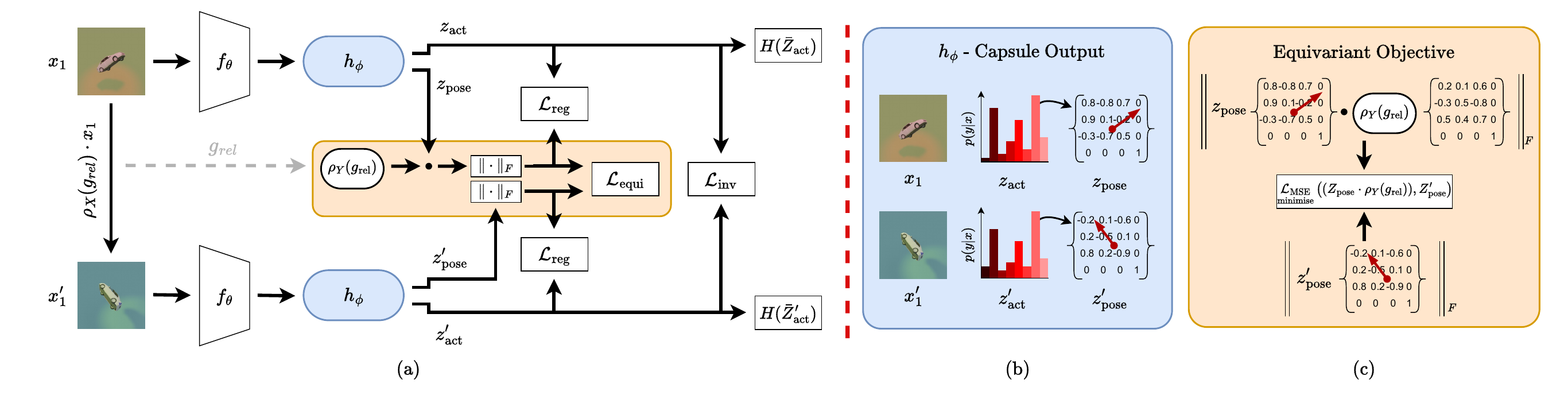}
    \caption{         \textbf{(a) Framework Overview.} Two views \({x}_1\) and \(x'_{1} = \rho_{X}(g_{rel}) \cdot x_{1}\) of the same object are processed by a shared CNN encoder \(f_{\theta}\) and a capsule-based projector \(h_{\phi}\).
        \textbf{(b) Projector's Output.} For each view, the projector produces an activation probability vector (\({z}_{{act}}\), \({z}'_{{act}}\)), indicating entity presence, and a pose matrix \(({z}_{{pose}}\), \({z}'_{{pose}}\)), 
        encoding the pose for each entity. 
        \textbf{(c) Equivariant Objective.} Equivariance is enforced by multiplying \({z}_{{pose}}\) by the relative transformation matrix \(\rho_Y(g_{\mathrm{rel}})\) and aligning the result with \({z}'_{{pose}}\).
     }
    \label{fig:overall}
\end{figure*}
\section{EquiCaps: Predictor-Free Pose-Aware SSL}
\label{sec:method}

\textbf{Motivation.} Prior equivariant self-supervised methods~\cite{park2022sen, equimod, SIE} commonly rely on predictor networks to enforce equivariant representations via augmentation-aware mappings. This leads to ad hoc vector representations that are both difficult to interpret or manipulate—an important consideration for evaluating equivariance. In effect, these methods attempt to enforce equivariance on architectures (i.e., CNNs) that are not naturally suited to it~\cite{capsules_survey} and often introduce extra complexity through dual projectors, representation splits, or hypernetworks.
Instead, we exploit the equivariant properties elicited by CapsNets~\cite{capsules_survey}, to more effectively encode positional information, particularly rotation and translation. Our design yields a simpler pose-aware SSL framework, outperforms previous equivariant methods on geometric tasks, and enables a more intuitive control and interpretability in the embedding space through the structured $4 \times 4$ capsule pose matrices. We introduce and rationalise each component of our methodology as follows.

\textbf{Framework Overview.} Our method adopts a Siamese joint embedding framework with two identical, weight-sharing branches. Each branch consists of a CNN $f_\theta$ followed by a CapsNet projector $h_{\phi}$. We omit global average pooling to preserve the 2D spatial structure necessary for capsule processing. Given an image $x \in \mathbb{R}^{c \times h \times w}$ from a dataset $D$, we sample two transformations \(g_1, g_2 \) from a distribution over the group ${G}$, as detailed in the supplementary Sec.~\ref{sec:dataset-generation}. These are applied to \(x\) to produce two distinct views, \( x_{g_1} = \rho_X(g_1) \cdot x \) and \( x_{g_2} = \rho_X(g_2) \cdot x \). Both views are passed through \(f_{\theta}\), producing feature maps \(F, F' \in \mathbb{R}^{c \times h \times w}\). These are then processed by \(h_\phi\), which contains \(N\) capsules, yielding \(z_{\mathrm{act}}, z_{\mathrm{pose}} = h_{\phi}(F)\) and \(z'_{\mathrm{act}}, z'_{\mathrm{pose}} = h_{\phi}(F')\), where \(z_{\mathrm{act}},\, z'_{\mathrm{act}} \in \mathbb{R}^{N}\) are the activation probabilities, and \(z_{\mathrm{pose}}, z'_{\mathrm{pose}} \in \mathbb{R}^{N \times 4 \times 4}\) are the corresponding pose matrices. The activation vectors encode transformation-invariant properties, representing semantic concepts that remain unchanged under the applied transformations \(g_1, g_2 \). In contrast, the pose matrices capture transformation-equivariant properties, encoding an entity's pose, where each capsule represents a unique entity. A visual depiction of the general architecture is given in~\cref{fig:overall}.

\textbf{Self-Routing CapsNet Projector.} 
To reduce the computational overhead of iterative-routing, we adopt the non-iterative self-routing algorithm~\cite{hahn2019self} as our projector. The projector \(h_{\phi}\) consists of two layers: a primary capsule layer that transforms encoder feature maps into poses (\(u_i\)) and activations (\(a_i\)),
and a self-routing layer, where each capsule independently determines its routing coefficients via a subordinate single-layer perceptron.  
To do this, the routing coefficients \(c_{ij}\) between each lower-layer capsule \(i\) and each upper-level capsule \(j\) are computed by multiplying the lower capsule’s pose vector \(u_i\) with a learnable weight matrix \(W^{\text{route}}\) and applying a softmax function.
The activation \(a_j\) of each upper-level capsule is then computed by forming votes from lower-level capsules—each vote is the product \(c_{ij}\, a_i\), which are then summed and normalised by the total lower-level activation, to formulate weighted votes: \vspace{-1ex}\begin{equation}
\begin{alignedat}{2}
c_{ij} &= \mathrm{softmax}\Bigl(W^{\text{route}}_{i}\, u_i\Bigr)_j, \qquad &
a_j &= \frac{\sum\nolimits_{i\in\Omega_l} c_{ij}\, a_i}{\sum\nolimits_{i\in\Omega_l} a_i}.
\end{alignedat}
\end{equation}
\noindent The capsule layer’s output pose is computed by multiplying a trainable weight matrix \(W^{\text{pose}}\) with each lower-level pose $u_i$, yielding a predicted pose $\hat{u}_{j|i}$ for each upper-level capsule. The final pose $u_j$ is the weighted average of $\hat{u}_{j|i}$:\vspace{-1ex}\begin{equation} 
\begin{alignedat}{2}
\hat{u}_{j|i} &= W^{\text{pose}}_{ij}\, u_i, \qquad &
u_j &= \frac{\sum\nolimits_{i\in\Omega_l} c_{ij}\, a_i\, \hat{u}_{j|i}}{\sum\nolimits_{i\in\Omega_l} c_{ij}\, a_i}.
\end{alignedat}
\end{equation}

\textbf{Enforcing Equivariance.} We define the relative transformation
\(g_{\mathrm{rel}} = g_2\,g_1^{-1},\) mapping between the views \(x_{g_1}\) and \(x_{g_2}\),
and let \(\rho_Y(g_{\mathrm{rel}})\) denote its matrix representation. To enforce equivariance (defined in~\cref{eq:equivariance}) we multiply \(z_{\mathrm{pose}}\) by \(\rho_Y(g_{\mathrm{rel}})\), aligning the transformed result with \(z_{\mathrm{pose}}'\). This indicates that the pose embeddings are equivariant w.r.t. the transformation group \(G\), ensuring that geometric changes in the input are consistently captured in the latent space. Each pose matrix \((z_{\mathrm{pose}} \cdot \rho_Y(g_{\mathrm{rel}}))\) and \(z_{\mathrm{pose}}'\) is normalised by its Frobenius norm and residing on a unit hypersphere. This direct manipulation in the latent space removes the need for a dedicated predictor module.

\textbf{Loss Functions.}
\noindent To enforce invariance we minimise the cross-entropy $H(Z_{\mathrm{act}}, Z'_{\mathrm{act}})$ between the activation probability vectors $Z_{\mathrm{act}}$ and $Z'_{\mathrm{act}}$, where $Z$ denotes the matrix embeddings computed over a batch. This aligns the capsule activations for different views of the same image. To ensure full capsule utilisation, we apply mean entropy maximisation (ME-MAX) regularisation~\cite{assran2021semi, assran2022masked, joulin2012convexrelaxationweaklysupervised} to the activation probabilities. Let 
\(
\bar{Z}_{\mathrm{act}} = \frac{1}{B}\sum_{i=1}^{B} Z_{\mathrm{act}},
\)
where \(B\) is the batch size. Maximising \(H(\bar{Z}_{\mathrm{act}})\) and \(H(\bar{Z}'_{\mathrm{act}})\) encourages a diverse distribution of activation across all capsules, preventing the collapse to only a few highly active capsules.

\noindent To enforce equivariance we minimise the mean-squared error between the normalised, transformed pose \((Z_{\mathrm{pose}} \cdot \rho_Y(g_{\mathrm{rel}}))\) and the corresponding pose \(Z'_{\mathrm{pose}}\).

\begin{table*}[ht!]
    \centering
    \caption{
    Evaluation on 3DIEBench and 3DIEBench-T using a ResNet-18 backbone. Learned representations are evaluated on an invariant task (classification) and equivariant tasks (rotation, translation, and colour prediction). For each group of methods, the best value is shown in bold. Supervised baselines are trained for each task. Equivariant methods are trained to be rotation-equivariant, with no explicit constraints for translation or colour prediction. 
    }
    \label{tab:allresults}
     \resizebox{0.8\textwidth}{!}{%
    \begin{tabular}{lccccccc}
    \toprule
    Method & \multicolumn{2}{c}{Classification (Top-1)} & \multicolumn{2}{c}{Rotation ($R^2$)} & \multicolumn{1}{c}{Translation ($R^2$)} & \multicolumn{2}{c}{Colour ($R^2$)} \\[0ex] 
    \cmidrule(lr){2-3} \cmidrule(lr){4-5} \cmidrule(lr){6-6} \cmidrule(lr){7-8} 
               & 3DIEBench & 3DIEBench-T & 3DIEBench & 3DIEBench-T & 3DIEBench-T & 3DIEBench & 3DIEBench-T \\
    \midrule
    \textcolor{gray}{\textit{Supervised Methods}} & & & & & & & \\
    ResNet-18 & 86.45 & 80.13 & 0.77 & 0.73 & 0.67 & 0.99 & 0.99 \\
    \midrule
    \textcolor{gray}{\textit{\shortstack{Invariant and Parameter \\
    Prediction Methods}}} & & & & & & & \\
    VICReg  & 84.28 & 74.71 & 0.45 & 0.39 & 0.22 & 0.10 & 0.50  \\
    SimCLR   & 86.73 & 80.08 & 0.52  & 0.44  & 0.25 & 0.29 & 0.50  \\
    SimCLR + AugSelf  & \textbf{87.44} & \textbf{80.86} & \textbf{0.75}  & \textbf{0.69}  &\textbf{0.50}  & 0.28 & 0.51 \\
    \midrule
    \textcolor{gray}{\textit{Equivariant Methods}} & & & & & & & \\
    SEN   & 86.99 & 80.20 & 0.51  & 0.45  & 0.26  & 0.29  & 0.47 \\
    EquiMod   & \textbf{87.39} & \textbf{80.76} & 0.50 & 0.43  & 0.24  & 0.29 & 0.38 \\
    SIE & 82.94 & 75.56 & 0.73 & 0.45 & 0.20  & 0.07 & 0.46 \\
    CapsIE & 79.14 & 75.20 & 0.74 & 0.60 & 0.46 & 0.04 & 0.03 \\
    \rowcolor{orange!15} 
    EquiCaps & 83.24 & 76.91 & \textbf{0.78} & \textbf{0.73} & \textbf{0.60} & 0.09 & 0.05 \\
    \bottomrule
    \end{tabular}%
    }
\end{table*}

To prevent collapse, 
we apply the variance-covariance regularisation criterion~\cite{bardes2022vicreg} to the concatenated capsule activation vectors and 
pose matrices, 
collectively denoted as \(Z_{\mathrm{cat}}\). Regularising these combined embeddings encourages diversity and non-redundancy across both activation and pose dimensions. 
The regularisation loss is defined as:
\begin{equation}
\mathcal{L}_{\mathrm{reg}}(Z_{\mathrm{cat}}) = \lambda_V\, V(Z_{\mathrm{cat}}) + \lambda_C\, C(Z_{\mathrm{cat}}),
\end{equation}\begin{equation}
V(Z_{\mathrm{cat}}) = \frac{1}{d} \sum_{j=1}^{d} \max\Bigl(0,\, 1 - \sqrt{\operatorname{Var}(Z_{\mathrm{cat}\cdot,j})}\Bigr)
\end{equation}\begin{equation}
C(Z_{\mathrm{cat}}) = \frac{1}{d} \sum_{i \neq j} \operatorname{Cov}(Z_{\mathrm{cat}})_{i,j}^2
\end{equation}
where $V(Z_{\mathrm{cat}})$ ensures that each dimension is utilised and $C(Z_{\mathrm{cat}})$ decorrelates different dimensions to reduce redundant information.
 
While SIE~\cite{SIE} and CapsIE~\cite{capsie} apply this regularisation before incorporating transformation parameters, we compute it directly on the transformation-aligned embedding. This approach yields improved performance in our experiments, as it operates on the final, fully aligned representation. 
Our overall loss function is defined as:
\begin{equation}
\begin{aligned}
\mathcal{L}
&= \lambda_{\mathrm{inv}}\, H(Z_{\mathrm{act}}, Z'_{\mathrm{act}}) + \\
&\quad  H(\bar{Z}_{\mathrm{act}}) + H(\bar{Z}'_{\mathrm{act}})  + 
\\
&\quad  \lambda_{\mathrm{equi}}\, \frac{1}{B}\sum_{i=1}^{B} \left\| \frac{Z_{i,\mathrm{pose}}
\cdot 
\rho_Y(g_{{\mathrm{rel},i}})} 
{\|Z_{i,\mathrm{pose}}
\cdot 
\rho_Y(g_{{\mathrm{rel},i}})\|_F} -
\frac{Z'_{i,\mathrm{pose}}}{\|Z'_{i,\mathrm{pose}}\|_F} \right\|_2^2 +\\
&\quad   L_{\mathrm{reg}}(Z_{\mathrm{cat}}) + L_{\mathrm{reg}}(Z'_{\mathrm{cat}}) 
\end{aligned}
\label{eq:final_loss}
\end{equation}
where \(\|\cdot\|_F\) denotes the Frobenius norm and \(
\lambda_{\text{inv}} = 0.1, \lambda_V = 10, \lambda_C = 1, \text{ and } 
 \lambda_{\text{equi}} = 5 
\). Finally, although the loss is calculated on the embeddings, the encoder’s globally pooled output is used for downstream tasks as is a practice shown to improve performance~\cite{chen2020simple}.

\textbf{Removal of Predictor Module.} 
\noindent Unlike previous equivariant methods such as SEN~\cite{park2022sen}, EquiMod~\cite{equimod}, SIE~\cite{SIE}, and CapsIE~\cite{capsie} that use a dedicated predictor network to map one embedding to another, our approach eliminates this module. 
This simplifies the architecture and instead enforces the explicit mapping of transformations in the embedding space via affine transformation matrices applied to the pose embeddings rather than manipulating embeddings through an intermediate parameter-prediction stage. 

\section{Experimental Results}
\label{sec:experiments}
\textbf{Setup.} To ensure fair comparisons, we adopt the experimental setup from~\cite{SIE} across all methods, which aligns with standard SSL practices: a pre-training phase followed by various task-specific downstream evaluations. A detailed description of the experimental setup is given in supplementary Sec.~\ref{sec:implementation-details}.

\textbf{Baselines.} As an upper bound, we include a supervised baseline using ResNet-18~\cite{resnet}.
For invariant self-supervision, we benchmark against VICReg~\cite{bardes2022vicreg}, a non-contrastive method, and SimCLR~\cite{chen2020simple}, a contrastive approach. We also consider SimCLR + AugSelf~\cite{lee2021improving}, a parameter-prediction extension of SimCLR that learns the augmentation parameters but does not enforce explicit equivariance. 
For equivariant self-supervision, we compare against SEN~\cite{park2022sen}, EquiMod~\cite{equimod}, SIE~\cite{SIE}, and CapsIE~\cite{capsie}. Following~\cite{SIE, stl}, we replace SEN’s triplet loss with the SimCLR loss to avoid hyperparameter tuning.

\textbf{Evaluation Protocol.}
For the invariant task, we perform classification via the linear evaluation protocol, where a linear classification layer is trained on the frozen representations.
To assess equivariant properties, we perform separate rotation and translation prediction tasks. For completeness, translation is evaluated in both the object and the base frame, with base frame results given in supplementary Sec.~\ref{sec:baseframe}.
For each task, a three-layer MLP is trained on the frozen representations. 
Its inputs are concatenated pairs of representations from two views of the same object, and is trained to regress the transformation parameters—rotation for the rotation task and translation for the translation task. 
To explore invariance to non-geometric transformations, a linear layer is trained on the frozen representations to regress the floor and spot hue. The input representations are identical to those used in the geometric tasks. Further details are provided in supplementary Sec.~\ref{sec:implementation-details}.

\subsection{Rotation-Only Pre-Training}\cref{tab:allresults} summarises the downstream performance of the learned representations on both 3DIEBench and 3DIEBench-T. Specifically, 3DIEBench evaluates classification, rotation, and colour (hue) tasks, while 3DIEBench-T adds translations.
Notably, all equivariant methods in~\cref{tab:allresults} are optimised solely for rotation equivariance during pre-training.
As observed in prior works, invariant methods fail to learn successful equivariant representations of the rotations and translations, while the addition of predictive tasks (AugSelf) improves their capability to quantify this change. 

We find that EquiCaps consistently outperforms SIE—its closest competitor—marginally on classification but significantly on geometric equivariant tasks, achieving $+$0.28 $R^2$ in rotation and $+$0.40 $R^2$ in translation on 3DIEBench-T. 
Notably, our capsule-based projector is able to intrinsically capture translational attributes, despite no direct supervision. CapsIE also generalises relatively better than other equivariant methods, but it underperforms our method, and validates our claim that its predictor-based design underuses the inherent pose-awareness capabilities of CapsNets.
Nevertheless, we find that non-capsule architectures do not generalise well under more complex, combined 3D transformations. We attribute this limitation to their lack of an explicit, pose-aware mechanism.

\subsection{Rotation-Translation Pre-Training} 
In~\cref{tab:3diesbenchtchange}, we present results when both rotation and translation parameters are optimised. As expected, all methods exhibit a small improvement in translation performance. 
However, the two capsule-based methods  
retain their substantial improvement on geometric tasks.
Consistent with~\cite{SIE_perform_drop_on_mult_transf_neurips_2024}, we find that SIE’s gains on 3DIEBench do not generalise well to multi-geometric learning. We further find that SIE exhibits minimal gains over purely invariant methods and maintains the weakest translation prediction, even when explicitly trained for translation equivariance.

\textbf{Colour Prediction.} 
EquiCaps maintains near-perfect invariance to hue transformations, consistent with prior work such as SIE and CapsIE, although SIE is less effective under 3DIEBench-T. 
By contrast, non-capsule methods exhibit substantially increased colour prediction when transitioning to 3DIEBench-T. We conjecture that the dataset’s increased geometric complexity makes the colour task easier to solve for models lacking explicit pose-aware mechanisms.

\subsection{Performance on Embeddings}\label{subsec:performance-embeddings}
As in prior works~\cite{chen2020simple, SIE}, we observe that representations learned by invariant methods achieve higher-than-expected performance on equivariant tasks. This arises because the invariance objective is enforced after the projector, while the representations can still retain information~\cite{chen2020simple}. To explore this further, we evaluate rotation and translation predictions on the embeddings (see~\cref{tab:rotation-embeddings3diebencht}), where all methods show decreased performance compared to their representation-level results, highlighting that the projector absorbs more invariance. We also observe that invariant methods, along with SEN and EquiMod, converge to similar embedding-level results, except for EquiMod, which retains higher translation prediction performance, as in the representation-stage results. SIE aligns with invariant methods on translation prediction, consistent with its representation stage. Notably, EquiCaps maintains the highest performance on equivariant tasks, and capsule-based methods suffer the smallest decline from representations to embeddings, underscoring the advantages of capsules' pose-awareness.

\begin{table} [h!]
    \centering
    \caption{Evaluation on 3DIEBench-T pre-trained for both rotation and translation. Learned representations are evaluated on an invariant and equivariant tasks: classification, rotation, translation, and colour prediction. We indicate the change relative to~\cref{tab:allresults}. }
    \resizebox{\columnwidth}{!}{%
    \begin{tabular}{lcccc}
    \toprule
    Method & Classification (Top-1) & Rotation ($R^2$) & Translation ($R^2$) & Colour ($R^2$) \\
    \midrule
    SimCLR + AugSelf & 81.04 $\uparrow$ 0.18 & 0.69 = 0.00 & 0.64 $\uparrow$ 0.14 & 0.51 = 0.00 \\ \midrule
    SEN     & 80.23 $\uparrow$ 0.03 & 0.46 $\uparrow$ 0.01 & 0.28 $\uparrow$ 0.02 & 0.50 $\uparrow$ 0.03\\
    EquiMod        & \textbf{80.89} $\uparrow$ 0.13 & 0.46 $\uparrow$ 0.03 & 0.37 $\uparrow$ 0.13 & 0.37 $\downarrow$ 0.01\\
    SIE            & 75.91 $\uparrow$ 0.35 & 0.48 $\uparrow$ 0.03 & 0.22 $\uparrow$ 0.02 & 0.36 $\downarrow$ 0.10\\
    CapsIE         & 76.31 $\uparrow$ 1.11 & 0.62 $\uparrow$ 0.02 & 0.53 $\uparrow$ 0.07 & 0.03 = 0.00\\
    \rowcolor{orange!15} 
    EquiCaps& 78.25 $\uparrow$ 1.34 & \textbf{0.71} $\downarrow$ 0.02 & \textbf{0.62} $\uparrow$ 0.02 & 0.02 $\downarrow$ 0.03\\
    \bottomrule
    \end{tabular}%
    }
    \label{tab:3diesbenchtchange}
\end{table}
\begin{table}[h!]
\centering

\caption{Evaluation on 3DIEBench and 3DIEBench-T pre-trained for both rotation and translation. Learned embeddings are evaluated on equivariant tasks: rotation and translation prediction.
}
\resizebox{1.0\columnwidth}{!}{%
\begin{tabular}{lccc}
\toprule
\multirow{2}{*}{Method} & \multicolumn{2}{c}{Rotation ($R^2$)} & \multicolumn{1}{c}{Translation ($R^2$)} \\
\cmidrule(lr){2-3}\cmidrule(lr){4-4}
 & 3DIEBench & 3DIEBench-T & 3DIEBench-T \\
\midrule
VICReg              & 0.24   & 0.24   & 0.00 \\
SimCLR              & 0.24   & 0.24   & 0.00 \\
SimCLR + AugSelf    & 0.24   & 0.24   & 0.00 \\\midrule
SEN                 & 0.24   & 0.24   & 0.00 \\
EquiMod             & 0.24   & 0.24   & 0.22 \\
SIE                 & 0.60   & 0.34   & 0.00 \\
CapsIE              & 0.71   & 0.58   & 0.26 \\
\rowcolor{orange!15} 
EquiCaps            & \textbf{0.73}   & \textbf{0.61}   & \textbf{0.33}  \\
\bottomrule
\end{tabular}%
}
\label{tab:rotation-embeddings3diebencht}
\end{table}

\begin{table*}[t]
    \centering
    \caption{
       Equivariance evaluation on 3DIEBench (top) and 3DIEBench-T (bottom), using
        MRR, H@k, and separate PRE measures for rotation and translation on 3DIEBench-T.
         We indicate which subset (train or val) is used to generate embeddings, and for PRE, which dataset (train, val, or all) is used for retrieval. For 3DIEBench-T, methods are pre-trained for both rotation and translation equivariance.
    }
    \label{tab:equiv_3diebench_combined}
    \resizebox{0.8\textwidth}{!}{%
    \begin{tabular}{lccccccccccccc}
        \toprule
         & \multicolumn{2}{c}{MRR ($\uparrow$)} 
         & \multicolumn{2}{c}{H@1 ($\uparrow$)} 
         & \multicolumn{2}{c}{H@5 ($\uparrow$)} 
         & \multicolumn{6}{c}{PRE ($\downarrow$)} \\\cmidrule(lr){2-3}
         \cmidrule(lr){4-5}\cmidrule(lr){6-7}\cmidrule(lr){8-13}
         & train & val
         & train & val
         & train & val
         & \multicolumn{2}{c}{train-train} 
         & \multicolumn{2}{c}{val-val}
         & \multicolumn{2}{c}{val-all} \\ \cmidrule(lr){2-3} \cmidrule(lr){4-5} \cmidrule(lr){6-7} 
         \cmidrule(lr){8-9}\cmidrule(lr){10-11}\cmidrule(lr){12-13}
         & All & All
         & All & All
         & All & All
         & Rotation & Tranlation
         & Rotation & Tranlation
         & Rotation & Tranlation \\
         \midrule
         \multicolumn{13}{c}{3DIEBench } \\
         \midrule

         SEN 
            & 0.09 & 0.09
            & 0.02 & 0.02
            & 0.10 & 0.10
            & 0.61 & --
            & 0.61 & --
            & 0.61 & -- \\
     EquiMod 
            & 0.09 & 0.09 
            & 0.02 & 0.02
            & 0.10 & 0.10
            & 0.61 & --
            & 0.61 & --
            & 0.61 & -- \\
         SIE 
            & 0.51 & 0.41
            & 0.41 & 0.30
            & 0.60 & 0.51
            & 0.26 & --
            & 0.29 & --
            & 0.27 & -- \\
         CapsIE 
            & 0.60 & 0.47
            & 0.50 & 0.36
            & 0.71 & 0.58
            & 0.17 & --
            & 0.21 & --
            & 0.20 & -- \\
            \rowcolor{orange!15} 
         EquiCaps 
            & \textbf{0.68} & \textbf{0.60}
            & \textbf{0.55} & \textbf{0.48}
            & \textbf{0.84} & \textbf{0.73}
            & \textbf{0.13} & --
            & \textbf{0.18} & --
            & \textbf{0.17} & -- \\
         \midrule
         \multicolumn{13}{c}{3DIEBench-T} \\
         \midrule
         SEN
            & 0.09 & 0.09
            & 0.02 & 0.02 
            & 0.10 & 0.10
            & 0.61 & 2.71
            & 0.62 & 2.72
            & 0.62 & 2.72 \\
         EquiMod
            & 0.13 & 0.12
            & 0.03 & 0.03
            & 0.18 & 0.15
            & 0.58 & 2.47
            & 0.59 & 2.51
            & 0.59 & 2.50 \\
         SIE
            & 0.15 & 0.14
            & 0.06 & 0.05
            & 0.20 & 0.18
            & 0.54 & 2.73
            & 0.54 & 2.72
            & 0.54 & 2.73 \\
         CapsIE
            & 0.46 & 0.37
            & 0.35 & 0.26
            & 0.55 & 0.47
            & \textbf{0.29} & 2.45
            & \textbf{0.31} & 2.48
            & \textbf{0.30} & 2.46 \\
            \rowcolor{orange!15} 
         EquiCaps
            & \textbf{0.59} & \textbf{0.51}
            & \textbf{0.48} & \textbf{0.41}
            & \textbf{0.70} & \textbf{0.62}
            & 0.30 & \textbf{2.34}
            & 0.35 & \textbf{2.41}
            & 0.33 & \textbf{2.38} \\
         \bottomrule
    \end{tabular}%
    }
\end{table*}

\begin{figure*}[b]
    \centering
     \captionsetup{skip=1pt}
     \includegraphics[width=0.9\linewidth]{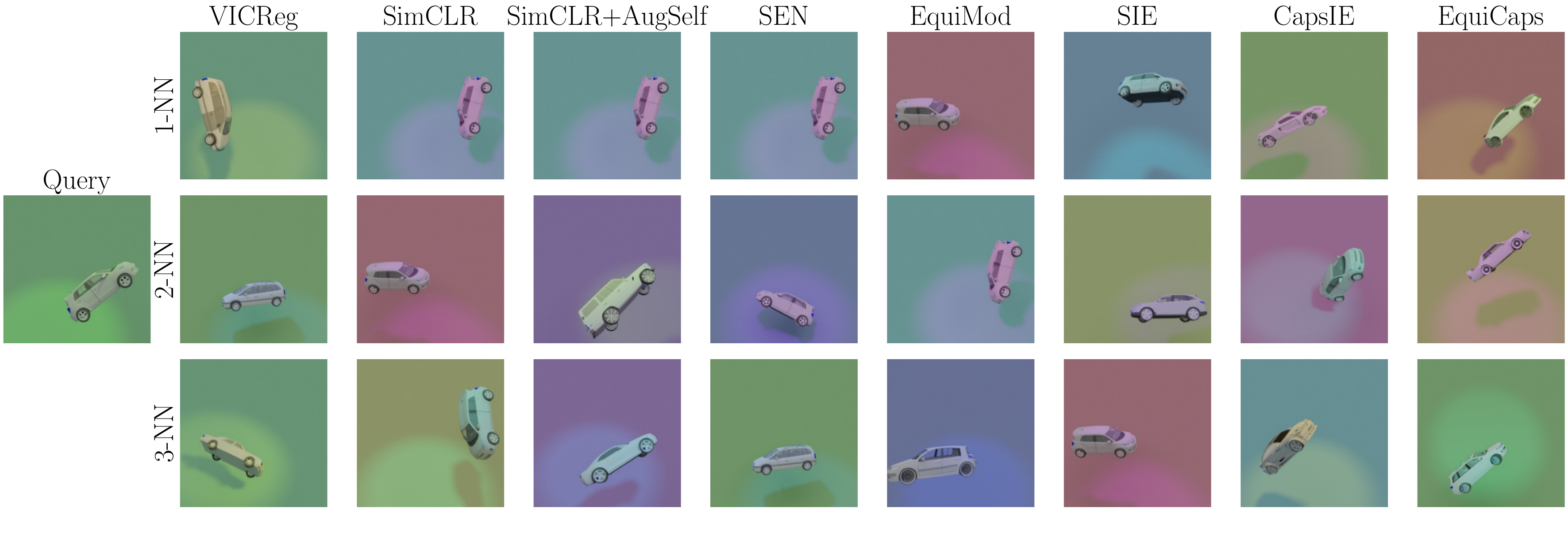}
    \caption{
    Nearest-neighbour representation retrieval on 3DIEBench-T validation set directly after pre-training. The query image (left) is compared against each method’s learned representations to find its top three nearest neighbours (in rows: 1-NN, 2-NN, 3-NN). Equivariant and parameter prediction methods are trained to be both rotationally and translationally equivariant. 
    }
    \label{fig:3diebencht_knn}
\end{figure*}
\newpage
\subsection{Equivariance Evaluation }
We employ and adapt the Mean Reciprocal Rank (MRR), Hit Rate at k (H@k), and Prediction Retrieval Error (PRE) metrics from~\cite{SIE}. For compared methods, for each object’s source and target poses, we feed the source embeddings and the corresponding transformation parameters through the predictor, obtaining a predicted embedding. To adapt it to our method, we multiply the projector’s pose matrix by the transformation matrix.
We obtain the nearest neighbours of the predicted embedding among the views of the same object. MRR is computed as the mean of the reciprocal rank $\frac{1}{r}$, where $r$ is the rank of the target embedding among the neighbours. H@k yields $1$ if the target is within the top-$k$ neighbours, and $0$ otherwise. PRE measures the mean distance between the transformation parameters of the nearest neighbour and the target across the dataset. 
Rotation uses $d_r = 1 - \langle q_1, q_2 \rangle^2$, where $q_1$ and $q_2$ are the nearest neighbour’s and target’s rotation represented as a quaternion, respectively.
Translation uses the mean-squared error $d_t = ||t_1 - t_2||_2^2$, where $t_1$ is the nearest neighbour’s translation vector and $t_2$ is the target’s translation vector.

As shown in~\cref{tab:equiv_3diebench_combined}, EquiCaps outperforms all methods across each data split of 3DIEBench, achieving an MRR of 0.60 on the validation set, well above SIE (0.41) and CapsIE (0.47). Similar results are observed in H@1/5 metrics, where SEN and EquiMod perform randomly (2\% and 10\%) on both data splits, while CapsIE’s PRE on all data splits approaches ours.
Although EquiCaps generally outperforms all methods on 3DIEBench-T across most metrics, CapsIE achieves slightly lower rotation PRE. 
Despite this, our method remains significantly stronger in MRR and H@1/5 across all data splits, with EquiMod slightly improving. In contrast, SIE's overall performance 
drops with H@1 of 0.05 on the validation set, marginally above random.

\subsection{Qualitative Results} 

\textbf{Nearest-Neighbour Representations.} We perform a nearest-neighbour representation search on the 3DIEBench-T validation set (\cref{fig:3diebencht_knn}) to illustrate the information captured by the representations. Invariant approaches should retrieve objects in the same class but varied poses, whereas equivariant methods should preserve the query pose. We find that only our method and CapsIE retrieve the nearest neighbour with a matching pose, though ours aligns also with the translation. For subsequent neighbours, only EquiCaps and AugSelf retrieve objects in similar poses, while CapsIE does so but less accurately. In contrast, the other methods, including equivariant methods, fail to yield pose-matching neighbours suggesting that invariance dominates and their representations leave out significant pose information. 
We replicate the retrieval on 3DIEBench (\cref{fig:3diebench_knn}, supplementary~\cref{sec:additionalqualres}), where ours, SIE, and CapsIE successfully return pose-matching neighbours, but remaining methods, only partially align poses, albeit less accurate. We conjecture this is attributed to the projector absorbing invariance, while partial rotation-related information remains at the representation stage.
All retrieved objects belong to the car category, indicating that class semantics are preserved consistent with our quantitative results.

\begin{figure}[t!]
    \centering
    \includegraphics[trim = 115 0 160 0 , clip, width=1\linewidth]{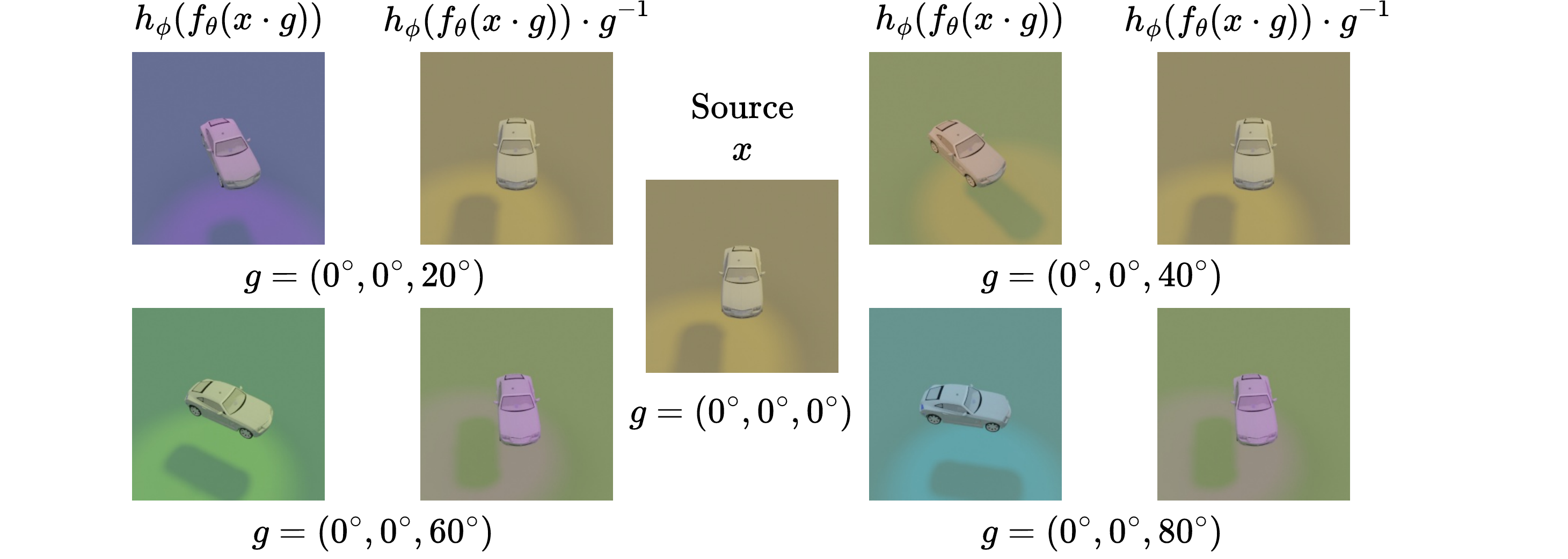}
    \caption{Illustration of transformation inversion of equivariant embeddings through capsule-based pose manipulation.}
    \label{fig:inverserot}
\end{figure}

\textbf{Equivariance via Pose Manipulation.} To further illustrate the equivariant properties of our approach, we examine the inversion property through the inverse rotation of an object in the embedding space. We rotate an object around the z-axis within the range $[0, \tfrac{\pi}{2}]$ in 5$^{\circ}$ increments and feed these images through our trained network to obtain its pose embedding. Embedding is subsequently multiplied by its corresponding inverse transformation matrix. To examine the impact of the inverse transformation we retrieve the nearest neighbour among all embeddings.
As shown in~\cref{fig:inverserot}, the nearest neighbour closely (within 5$^{\circ}$ error) or exactly matches the source view, highlighting the invertibility of our representations, and the ability to intuitively and directly steer the latent space. Further qualitative results are provided in supplementary Sec.~\ref{sec:additionalqualres}.

\subsection{Sensitivity Analysis \& Ablation Studies}
We conduct further analyses, including the impact of the number of capsules, invariant components, and experiments using a Vision Transformer backbone. Complete results and their findings are given in supplementary Sec.~\ref{sec:ablations} and ~\ref{sec:vit}. 

\begin{figure}
    \centering
    \includegraphics[width=1\linewidth]{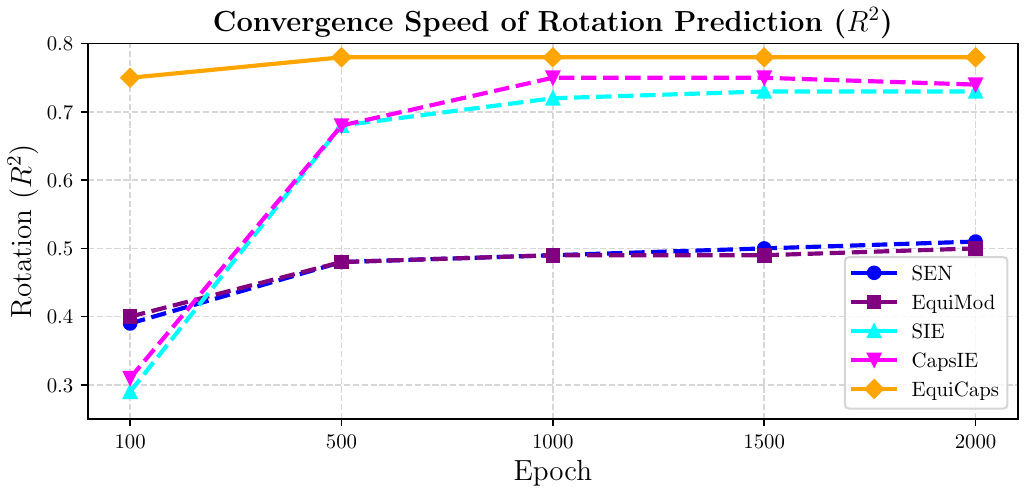}
        \caption{Convergence speed in epochs of Equivariant SSL methods on 3DIEBench rotation prediction.}
    \label{fig:epochsrotfig}
\end{figure}

\textbf{Convergence Speed.} \noindent The effectiveness of CapsNets intrinsic pose-aware ability is uniquely demonstrated by ~\cref{fig:epochsrotfig}. We show that on 3DIEBench, EquiCaps converges the fastest in rotation prediction, reaching $R^2=$ 0.75 by epoch 100 and stabilising at $R^2=$ 0.78 by epoch 500. On 3DIEBench-T, it remains the quickest, and a similar pattern emerges for classification on 3DIEBench, though some methods exhibit competitive convergence in classification performance on 3DIEBench-T.
Overall, our method achieves similar results while requiring fewer training epochs and, consequently, less compute.

\section{Conclusion}
\label{sec:conclusion}

We propose EquiCaps, a capsule-based SSL method that jointly learns invariant and equivariant representations. By leveraging capsules’ innate pose-awareness, EquiCaps shapes the latent space predictably, removing the need for a dedicated predictor. To enable richer benchmarking, we extend 3DIEBench to 3DIEBench-T, incorporating object translations alongside rotations. Empirically, EquiCaps achieves state-of-the-art rotation and object frame translation prediction among equivariant methods. Tests on 3DIEBench-T highlight capsule architectures’ superior generalisation under combined geometric transformations, demonstrating their potential in invariant-equivariant SSL.

\noindent \textbf{Limitations and Future Work.} Joint invariance-equivariance optimisation involves trade-offs; while EquiCaps balances these, task-specific prioritisation may vary. The method relies on known group elements—a constraint shared by existing methods—limiting applicability without such data, though partial information suffices. 
EquiCaps can theoretically handle any transformation 
which can be expressed as a matrix without architectural changes.
Future work includes other 3D transformations (e.g., scaling, shear), and tasks like object detection.

\section{Acknowledgements}
This work was supported by the UKRI AI Centre for Doctoral Training project SUSTAIN [grant reference: EP/Y03063X/1]. We also thank the University of Aberdeen's HPC Maxwell facility.

{
    \small
    \bibliographystyle{ieeenat_fullname}
    \bibliography{main}
}

\clearpage
\appendix 
\renewcommand{\thefigure}{A.\arabic{figure}}
\renewcommand{\thetable}{A.\arabic{table}}
\renewcommand{\theequation}{A.\arabic{equation}}

\section{Implementation Details}\label{sec:implementation-details}

\subsection{Reproducibility}

All pre-training experiments employed three NVIDIA A100 80 GB GPUs and took approximately 24 hours.
The code, dataset, and weights are released.
\subsection{Pre-training}

All methods use ResNet-18~\cite{resnet} 
as the base encoder network. For the compared methods except CapsIE, the projector is a three layer MLP. For capsule-based methods, the projector consists of 32 capsules. Training lasts 2000 epochs with a batch size of 1024 to ensure convergence. The Adam optimiser~\cite{adamoptimizer}
is employed with a learning rate of $10^{-3}$, $\beta_1 = 0.9$, $\beta_2 = 0.999$, and a weight decay of $10^{-6}$. 
We mention the hyperparameters of each method below.

  \textbf{Supervised ResNet-18}
The training and evaluation protocols are identical to those of the self-supervised setups.

  \textbf{VICReg} The projector is configured with intermediate dimensions of 2048-2048-2048, with loss weights 
$\lambda_{\text{inv}} = $ $\lambda_{\text{V}} = 10$, and   $\lambda_C = 1$.

  \textbf{SimCLR, SEN}
The projector is configured with intermediate dimensions of 2048-2048-2048, and the temperature parameter of the loss is set to 1. 

  \textbf{SimCLR + AugSelf }
 The projector is configured with intermediate dimensions of 2048-2048-2048, and the temperature parameter of the loss is set to 1. The parameter prediction head is configured with a MLP with intermediate dimensions 1024-1024-4. The two losses, SimCLR and parameter prediction, are assigned equal weight.

 \textbf{EquiMod}
In alignment with the original protocol, the projector is configured with intermediate dimensions of 1024-1024-128, 
The temperature parameter of the loss is set to 0.1. The two losses, invariance and equivariance, are assigned equal weight.

  \textbf{SIE}
Aligning with the original protocol, both of the invariance and equivariant projector are configured with intermediate dimensions 1024-1024-1024, and the loss weights $\lambda_{\text{inv}} = \lambda_V = 10$, $\lambda_{\text{equi}} = 4.5$, and $\lambda_C = 1$.
 
  \textbf{CapsIE, EquiCaps}
The projector is configured with 32 capsules and the loss weights $\lambda_{\text{inv}} = 0.1$, $\lambda_{\text{equi}} = 5$, $\lambda_V = 10$, and $\lambda_C = 1$.

\newpage
\subsection{Evaluation} 

  \textbf{Semantic Classification} A linear classification layer is trained on the frozen representations. The Adam optimiser is used with a learning rate of $10^{-3}$, $\beta_1 = 0.9$, $\beta_2 = 0.999$, and no weight decay. Training is carried out for 300 epochs with a batch size of 256, using the cross-entropy loss function. Performance is evaluated on the validation set comprising objects excluded from the training set.

  \textbf{Rotation Prediction} A three-layer MLP with intermediate dimensions 1024-1024-4 and intermediate ReLU activations is trained on the frozen representations. The inputs are concatenated pairs of representations from two distinct views of an object. The MLP is trained to regress the rotation between these views. The Adam optimiser is used with a learning rate of $10^{-3}$, $\beta_1 = 0.9$, $\beta_2 = 0.999$, and no weight decay. Training is conducted for 300 epochs with a batch size of 256, using the mean-squared error loss function. Performance is evaluated on the validation set using  \(R^2\), which contains objects excluded from the training set, with the object rotation range the same across both sets.

  To evaluate the equivariance performance, the metric \(
R^2 = 1 - \frac{\sum_i (y_i - \hat{y}_i)^2}{\sum_i (y_i - \bar{y})^2}
\) is used,
where 
\(\{y_i\}\) are the target values, \(\bar{y}\) is their mean, and \(\{\hat{y}_i\}\) are the predictions. Higher \(R^2\) values indicate a better fit for the predictions.

  \textbf{Translation Prediction }
The same methodology as for rotation prediction is followed. The only modification is that the output dimension of the MLP is three, corresponding to the elements of the translation vector, since is trained to regress the translation between the selected views.  

 \textbf{Colour Prediction} 
A linear layer is trained on top of frozen representations to regress the floor and spot hue. 
The inputs are concatenated pairs of representations from two distinct views of an object. The Adam optimiser is used with a learning rate of $10^{-3}$, $\beta_1 = 0.9$, $\beta_2 = 0.999$, and no weight decay. Training is conducted for 50 epochs with a batch size of 256, using the mean-squared error loss function. Performance is evaluated on the validation set using  \(R^2\).

\newpage

\section{Sensitivity Analysis \& Ablation Studies}\label{sec:ablations}
\subsection{Convergence Speed}

We examine convergence speed on 3DIEBench and 3DIEBench-T. As shown in~\cref{tab:rotation-epochs} for rotation prediction on 3DIEBench, our method converges fastest, reaching an $R^2$ of 0.75 by 100 epochs and stabilising at 0.78 by 500 epochs, significantly ahead of the other methods and more than double the early-stage performance of SIE and CapsIE. As shown in~\cref{tab:classification-epochs}, a similar trend emerges in classification, where our method achieves the highest accuracy at 100 epochs—surpassing even the purely invariant methods. While longer training further improves our classification performance, our approach gains less from 500 to 2000 epochs compared to the rest methods.

Convergence-speed results on 3DIEBench-T for rotation prediction and classification appear in~\cref{tab:rotation-epochs3diebencht} and~\cref{tab:classification-epochs3diebencht}, respectively. As with the rotation prediction task on 3DIEBench, our method still converges the fastest, though its relative convergence speed between 100 and 500 epochs is slower. In classification, other methods exhibit comparable performance. We attribute this to the added complexity of 3DIEBench-T as learning more complex geometric transformations demands more epochs for EquiCaps to encode all transformations.

\begin{table}[h]
\centering
\caption{Impact of training duration on 3DIEBench rotation prediction performance on learned representations.}
\resizebox{\columnwidth}{!}{%
\begin{tabular}{lccccc}
\toprule
 & \multicolumn{5}{c}{Rotation ($R^2$)} \\
\cmidrule(lr){2-6}   
Method & 100 ep. & 500 ep. & 1000 ep. & 1500 ep. & 2000 ep. \\
\midrule
VICReg & 0.28 & 0.46 & 0.46  & 0.46 & 0.45 \\
SimCLR & 0.42 & 0.48 & 0.49  & 0.51 & 0.52 \\
SimCLR + AugSelf & 0.50 & 0.71 & 0.74  & 0.75 & 0.75 \\
SEN & 0.39 & 0.48 & 0.49  & 0.50 & 0.51 \\
EquiMod & 0.40 & 0.48 & 0.49  & 0.49 & 0.50 \\
SIE & 0.29 & 0.68 & 0.72  & 0.73 & 0.73 \\
CapsIE & 0.31 & 0.68 & 0.75  & 0.75 & 0.74 \\
\rowcolor{orange!15}  
EquiCaps & 0.75 & 0.78 & 0.78  & 0.78 & 0.78 \\
\bottomrule
\end{tabular}%
}
\label{tab:rotation-epochs}
\end{table}

\begin{table} [h]
\centering
\caption{Impact of training duration on 3DIEBench classification performance on learned representations.}
\resizebox{\columnwidth}{!}{%
\begin{tabular}{lccccc}
\toprule
 & \multicolumn{5}{c}{Classification (Top-1)} \\
\cmidrule(lr){2-6}   
Method & 100 ep. & 500 ep. & 1000 ep. & 1500 ep. & 2000 ep. \\
\midrule
VICReg & 49.12 & 79.40 & 83.12  & 84.43 & 84.28 \\
SimCLR & 72.58 & 84.00 & 85.87  & 86.61 & 86.73 \\
SimCLR + AugSelf & 73.50 & 84.57 & 86.51  & 87.08 & 87.44 \\
SEN  & 67.24 & 82.36 & 85.29  & 86.45 & 86.99 \\  
EquiMod    & 73.19 & 84.89 & 86.36  & 87.00 & 87.39 \\
SIE             & 51.49 & 77.59 & 81.05  & 82.12 & 82.94 \\
CapsIE  & 46.12 & 72.60 & 78.68  & 79.54 & 79.14 \\
\rowcolor{orange!15}  
EquiCaps & 75.44 & 81.10 & 82.36 & 82.82 & 83.24 \\
\bottomrule
\end{tabular}%
}
\label{tab:classification-epochs}
\end{table}
\newpage

\begin{table}[h]
\centering
\caption{Impact of training duration on 3DIEBench-T rotation performance on learned representations. Equivariant methods are pre-trained for both rotation and translation.}
\resizebox{\columnwidth}{!}{%
\begin{tabular}{lccccc}
\toprule
 & \multicolumn{5}{c}{Rotation ($R^2$)} \\
\cmidrule(lr){2-6}   
Method & 100 ep. & 500 ep. & 1000 ep. & 1500 ep. & 2000 ep. \\
\midrule
VICReg  & 0.25 & 0.24& 0.25 &0.36 &   0.39\\
SimCLR & 0.37 & 0.44& 0.44 &0.45 & 0.44 \\
SimCLR + AugSelf  & 0.41 &0.63 & 0.67 & 0.68&0.69  \\
SEN  & 0.34 & 0.46&0.46 &0.46 & 0.46 \\
EquiMod  &0.38  & 0.46& 0.46 &0.45 &0.46  \\
SIE  & 0.26 & 0.44 & 0.47 & 0.48 & 0.48 \\
CapsIE  & 0.25 & 0.49 & 0.62 & 0.62 &  0.62  \\
\rowcolor{orange!15}  
EquiCaps & 0.54 & 0.70 & 0.71 & 0.70 & 0.71 \\
\bottomrule
\end{tabular}%
}
\label{tab:rotation-epochs3diebencht}
\end{table}

\begin{table} [h]
\centering
\caption{Impact of training duration on 3DIEBench-T classification performance on learned representations. Equivariant methods are pre-trained for both rotation
and translation.}
\resizebox{\columnwidth}{!}{%
\begin{tabular}{lccccc}
\toprule
 & \multicolumn{5}{c}{Classification (Top-1)} \\
\cmidrule(lr){2-6}   
Method & 100 ep. & 500 ep. & 1000 ep. & 1500 ep. & 2000 ep. \\
\midrule
VICReg & 31.50 & 19.46& 31.80 &66.60 &74.71 \\
SimCLR& 67.56 &78.32 &79.52  &79.87 & 80.08\\
SimCLR + AugSelf &  66.02& 78.80&80.04  &80.77 &81.04 \\
SEN & 59.64 & 77.11& 79.17 & 79.66& 80.23\\
EquiMod    & 69.02 & 79.31&80.38 &80.85 & 80.89\\
SIE           & 44.97 &67.72 & 73.92& 75.58  &75.91 \\
CapsIE & 37.22 & 65.14& 74.13 & 75.63& 76.31\\
\rowcolor{orange!15}  
EquiCaps & 62.00 & 75.87 & 77.31  & 77.45 & 78.25\\
\bottomrule
\end{tabular}%
}
\label{tab:classification-epochs3diebencht}
\end{table}

\subsection{Number of Capsules} 
We report in~\cref{tab:num_capsules} that increasing the number of capsules improves classification and rotation prediction on 3DIEBench. 
A similar trend appears for 3DIEBench-T, though geometric tasks depend on explicit optimisation.
Specifically, from 32 to 64 capsules yields a slight decrease in translation when optimising only for rotation, suggesting that the additional capacity is primarily dedicated to the explicit rotation objective. Nonetheless, when we optimise for both objectives, translation performance increases substantially, confirming the effectiveness of the $4 \times 4$ capsule pose structure and our proposed matrix manipulation while 
classification, rotation and translation performance remain near supervised level. We also observe that as the number of capsules grows, the network tends to focus more on geometric tasks relative to colour prediction.

\begin{table*}[]
\centering
\caption{
Impact of the number of capsules in the EquiCaps projector on 3DIEBench and 3DIEBench-T. \textsuperscript{†}Denotes pre-trained for both rotation and translation.
We evaluate invariance via classification and equivariance via rotation, translation, and colour prediction tasks.
}

\resizebox{1\textwidth}{!}{%
\begin{tabular}{l c c c c c c c c c c c}
\toprule
No. of Capsules & \multicolumn{3}{c}{Classification (Top-1)} & \multicolumn{3}{c}{Rotation ($R^2$)} & \multicolumn{2}{c}{Translation ($R^2$)} & \multicolumn{3}{c}{Colour ($R^2$)} \\
\cmidrule(lr){2-4} \cmidrule(lr){5-7} \cmidrule(lr){8-9} \cmidrule(lr){10-12}
 & 3DIEBench & \multicolumn{2}{c}{3DIEBench-T} & 3DIEBench & \multicolumn{2}{c}{3DIEBench-T} & \multicolumn{2}{c}{3DIEBench-T} & 3DIEBench & \multicolumn{2}{c}{3DIEBench-T} \\
\midrule
16 & 81.89 & 73.86 & 74.13\textsuperscript{†} & 0.77 & 0.71 & 0.67\textsuperscript{†} &
\textbf{ } 
0.57 &  0.56\textsuperscript{†}& 0.13 & 0.04 &0.04\textsuperscript{†}  \\
32 & 83.24 & 76.91 & 77.88\textsuperscript{†} & 0.78 & 0.73 & 0.71\textsuperscript{†} & \textbf{ } 
0.60 & 0.61\textsuperscript{†} & 0.09 & 0.05 &0.02\textsuperscript{†}  \\
64 & 83.66 & 77.96 & 78.80\textsuperscript{†} & 0.79 & 0.74 & 0.71\textsuperscript{†} & 
\textbf{ } 
0.53 & 0.64\textsuperscript{†} & 0.05 & 0.01 &  0.01\textsuperscript{†}\\
\bottomrule
\end{tabular}%
}
\label{tab:num_capsules}
\end{table*}
\newpage
\subsection{Invariance Loss Function: Components}
\begin{table*}
\centering
\caption{Ablation study of different loss function components contributing to invariance in EquiCaps, evaluated on 3DIEBench downstream tasks. We evaluate learned representations on invariance (classification) and equivariance (rotation and colour prediction via \(R^2\)). The losses considered are \(\mathcal{L}_{\mathrm{inv}}\) (invariance), 
\(\mathcal{L}_{\mathrm{ME\textrm{-}MAX}}\) (mean entropy maximisation regularisation), 
\(\mathcal{L}_{\mathrm{equi}}\) (equivariance), 
and \(\mathcal{L}_{\mathrm{reg}}\) (variance-covariance regularisation). \(\mathcal{L}_{\mathrm{reg}}\) can be applied either to 
${Z}_{\mathrm{cat}}$, concatenating both activation and pose matrices, or to ${Z}_{\mathrm{pose}}$, which is applied only to the pose matrices.} 
\resizebox{\textwidth}{!}{%
\begin{tabular}{lccccccc}
\toprule
 & \multicolumn{4}{c}{{Loss Functions}} & {Classification (Top-1)} & {Rotation (R$^2$)} & {Colour (R$^2$)} \\
\cmidrule(lr){2-5} \cmidrule(lr){6-6} \cmidrule(lr){7-7} \cmidrule(lr){8-8}
{Method} &  {$\mathcal{L}_{\mathrm{inv}}$} & {$\mathcal{L}_{\mathrm{ME-MAX}}$} & \textbf{$\mathcal{L}_{\mathrm{equi}}$} & {$\mathcal{L}_{\mathrm{reg}}$} & & & \\
\midrule
EquiCaps OnlyEqui & - & - & \checkmark & ${Z}_{\mathrm{pose}}$ & 81.68 & 0.78 & 0.01 \\
EquiCaps w/o {$\mathcal{L}_{\mathrm{ME-MAX}}$} \& {$\mathcal{L}_{\mathrm{inv}}$} & - & - & \checkmark & ${Z}_{\mathrm{cat}}$ & 82.40 & 0.78 & 0.04\\
EquiCaps OnlyEqui w/ {$\mathcal{L}_{\mathrm{ME-MAX}}$} & - & \checkmark & \checkmark & ${Z}_{\mathrm{pose}}$ & 81.61 & 0.78 & 0.06\\
EquiCaps w/o {$\mathcal{L}_{\mathrm{inv}}$} & - & \checkmark & \checkmark & ${Z}_{\mathrm{cat}}$ & 81.43 & 0.78 & 0.06\\ \midrule
\rowcolor{orange!15}  
EquiCaps & \checkmark & \checkmark & \checkmark & ${Z}_{\mathrm{cat}}$ & 83.24 & 0.78 & 0.09 \\ 
\bottomrule
\end{tabular}}
\label{tab:lossfn-ablation}
\end{table*}

We examine how different invariance-related loss function components influence the performance of EquiCaps on downstream tasks using the 3DIEBench. Specifically, we evaluate the impact on classification (invariance), rotation and colour prediction (equivariance), on learned representations. We investigate the following configurations:

\begin{enumerate}[label=(\alph*)]
    \item \(\mathcal{L}_{\mathrm{equi}} + \mathcal{L}_{\mathrm{reg}}\bigl(Z_{\mathrm{pose}}\bigr)\): 
    Equivariance combined with variance-covariance regularisation applied only to the pose matrices. 
    
    \item \(\mathcal{L}_{\mathrm{equi}} + \mathcal{L}_{\mathrm{reg}}\bigl(Z_{\mathrm{cat}}\bigr)\): 
    Equivariance combined with variance-covariance regularisation applied on the concatenated activation vectors and pose matrices.
    
    \item \(\mathcal{L}_{\mathrm{equi}} + \mathcal{L}_{\mathrm{reg}}\bigl(Z_{\mathrm{pose}}\bigr) + \mathcal{L}_{\mathrm{ME\text{-}MAX}}\): 
    As in (a), but combined with mean-entropy maximisation regularisation (\(\mathcal{L}_{\mathrm{ME\text{-}MAX}}\)).
    
    \item \(\mathcal{L}_{\mathrm{equi}} + \mathcal{L}_{\mathrm{reg}}\bigl(Z_{\mathrm{cat}}\bigr) + \mathcal{L}_{\mathrm{ME\text{-}MAX}}\):
    As in (b), but combined with \(\mathcal{L}_{\mathrm{ME\text{-}MAX}}\).
    
    \item Our EquiCaps method as defined in~\cref{eq:final_loss} that combines invariance, equivariance, variance-covariance regularisation applied on the concatenated activation vectors and pose matrices, and mean-entropy maximisation regularisation.
\end{enumerate}

We observe in~\cref{tab:lossfn-ablation} that including all of our invariance-related losses {$\mathcal{L}_{\mathrm{inv}}$}, \(\mathcal{L}_{\mathrm{ME\text{-}MAX}}\), \(    \mathcal{L}_{\mathrm{reg}}\bigl(Z_{\mathrm{cat}}\bigr) \),  combined with \(\mathcal{L}_{\mathrm{equi}}  \) improve classification without compromising rotation performance, and we maintain almost perfect invariance to colour hue transformations. As expected, the mean-entropy maximisation \(\mathcal{L}_{\mathrm{ME\text{-}MAX}}\) alone does not suffice to boost classification in the absence of our invariance loss, regardless of whether the variance-covariance regularisation is applied to the pose matrices or to the concatenated embeddings. This finding suggests that  \(\mathcal{L}_{\mathrm{ME\text{-}MAX}}\) most effectively distributes activations when combined with the invariance loss. Similarly, whether we apply the variance-covariance regularisation solely to the pose matrices or to the concatenated embeddings, including the invariance loss and \(\mathcal{L}_{\mathrm{ME\text{-}MAX}}\) still yields further gains in classification. 
We also find that in every ablation setting, rotation performance remains stable and on par with the supervised baseline. 
Nonetheless, our proposed method further enhances semantic representation while retaining almost perfect invariance to colour.

\section{Additional Qualitative Results}\label{sec:additionalqualres}
\subsection{Nearest-Neighbour Representations on 3DIEBench}

\begin{figure*}
 \caption{Nearest-neighbour representation retrieval on 3DIEBench validation set directly after pre-training. The query image (left) is compared against each method’s learned representations to find its top three nearest neighbours (in rows: 1-NN, 2-NN, 3-NN).  }
    \centering
     \captionsetup{skip=1pt}
     \includegraphics[width=\linewidth]{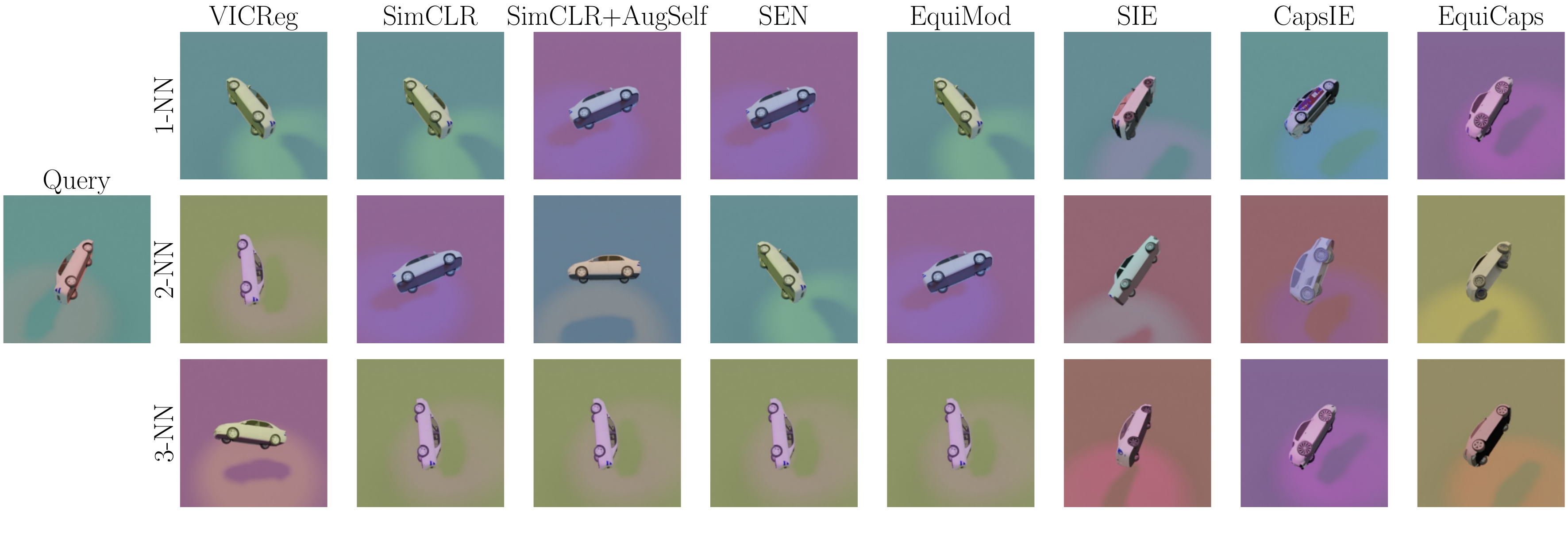}
   
    \label{fig:3diebench_knn}
\end{figure*}
We replicate the retrieval of nearest representations from \cref{fig:3diebencht_knn} performed on the 3DIEBench-T dataset, on 3DIEBench and show the results in \cref{fig:3diebench_knn}. We observe that our method, CapsIE, and SIE consistently retrieve nearest neighbours in similar poses as with the query, consistent with their high quantitative rotation prediction results. For the remaining methods, we observe that the learned invariance dominates, yielding mostly a range of object poses among the retrieved samples. However, aside from VICReg which showed the highest level of invariance in the quantitative results, these methods also retrieve some nearest neighbours in similar poses—albeit less accurately—demonstrating that partial rotation-related information remains at the representation stage. 
Overall, these findings are consistent with our quantitative results.

\begin{figure}[h]
    \centering    \caption{Illustration of equivariant capsule-based pose manipulation. We observe that our pose embeddings change predictably based on the applied transformation.}
    \includegraphics[width=1\linewidth]{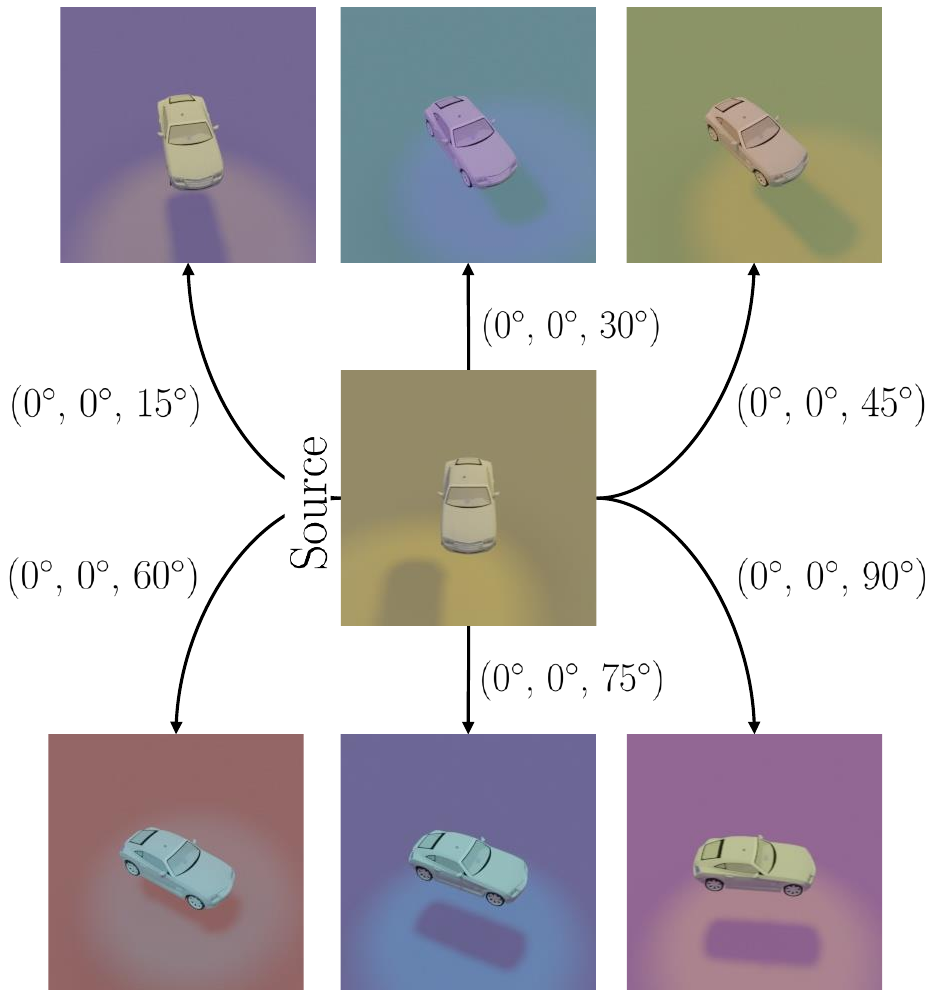}

    \label{fig:posemanipulation2}
\end{figure}
\subsection{Equivariance via Pose Manipulation}

In addition to the qualitative results shown in~\cref{fig:inverserot}, we further illustrate our method’s equivariant properties by performing the inverse of the previous experiment. Specifically, we rotate an object around the z-axis within the range $[0, \tfrac{\pi}{2}]$ in 5$^{\circ}$ increments, and feed each of these generated and original (source) pose into our projector to obtain its pose embeddings. Next, instead of transforming each embedding by the inverse rotation, we multiply the source pose embedding by the corresponding transformation matrix as shown in~\cref{fig:posemanipulation2}. For each transformed embedding we retrieve the nearest neighbour among all embeddings. Observing how each nearest neighbour closely changes according to the applied latent transformation further highlights our methods’ equivariant properties and its capability to preserve and manipulate pose information directly in the latent space. 
\section{Vision Transformer Backbone Evaluation}\label{sec:vit}
We re-implement the best-performing equivariant models with a ViT-Tiny/16 \citep{caron2021emerging} backbone using a patch size of 16. Training, validation and hyperparameters are identical to the ResNet-18 setting.
For capsule-based methods, given the [CLS] token and $N$ patch embeddings, ${z}\in\mathbb{R}^{(N+1)\times D}$, produced by the ViT encoder, we discard the [CLS] token and reshape the remaining $N$ patch embeddings into ${z_p}\in\mathbb{R}^{D\times H'\times W'}$ to recover the 2D spatial structure needed for capsule processing, where $N$ is the number of patches, $D$ is the embedding dimension, and $H'=W'=\sqrt{N}$. For non-capsule-based methods, after discarding the [CLS] token, we average the patch embeddings.

\cref{tab:3diebench_comparison} shows that EquiCaps with a ViT backbone performs similarly in classification accuracy compared to the supervised ViT, while there is a marginal decrease in geometric equivariance. EquiCaps consistently outperforms both SIE and CapsIE, which drop in classification performance and largely fail on the geometric tasks when moved from CNNs to ViTs.
On colour prediction, EquiCaps remains largely invariant to background hue transformations, whereas the rest methods degrade.
Overall, EquiCaps is the most robust equivariant method across all metrics, even on a ViT backbone.

\begin{table*}
\centering
\caption{Evaluation on 3DIEBench and 3DIEBench-T using a ViT-Tiny/16 backbone.  
Learned representations are evaluated on an invariant task (classification) and three
equivariant tasks (rotation, translation and colour).  
Supervised baselines are trained per task.  Equivariant methods are optimised for rotation only.}
\label{tab:3diebench_comparison}
\resizebox{1\textwidth}{!}{%
\begin{tabular}{l c c c c c c c}
\toprule
Method & \multicolumn{2}{c}{Classification (Top-1)} 
      & \multicolumn{2}{c}{Rotation ($R^{2}$)} 
      & \multicolumn{1}{c}{Translation ($R^{2}$)} 
      & \multicolumn{2}{c}{Colour ($R^{2}$)} \\

\cmidrule(lr){2-3} \cmidrule(lr){4-5} \cmidrule(lr){6-6} \cmidrule(lr){7-8}
      & 3DIEBench & 3DIEBench-T 
      & 3DIEBench & 3DIEBench-T 
      & 3DIEBench-T
      & 3DIEBench & 3DIEBench-T \\ 
\midrule
\textcolor{gray}{\textit{Supervised Methods}} \\
ViT-Tiny/16 & 84.70 & 76.57 & 0.78 & 0.73 & 0.68 & 0.99 & 0.99 \\ \midrule
\textcolor{gray}{\textit{Equivariant Methods}}\\
SIE        & 78.49 & 62.97 & 0.39 & 0.30 & 0.11 & 0.23 & 0.38 \\
CapsIE     &     61.80   &    39.01    &  0.33    &   0.26   &  0.10  & 0.31     &     0.61 \\
\rowcolor{orange!15}
EquiCaps   & \textbf{84.14} & \textbf{76.25} 
           & \textbf{0.71} & \textbf{0.69} 
           & \textbf{0.50} 
           & 0.06 & 0.13 \\
\bottomrule
\end{tabular}%
}
\end{table*}

\begin{table*}[b]
    \centering
    \caption{Evaluation on 3DIEBench-T using a ResNet-18 backbone. Learned representations are evaluated on an invariant task (classification) and equivariant tasks (rotation, translation, and colour prediction). For each group of methods, the best value is shown in bold. 
}
    \label{tab:comparison-full}
    \begin{subtable}[t]{0.49\textwidth}
        \centering
        \caption{Rotation-only Pre-training. Equivariant methods are trained to be rotation-equivariant only.}
        \label{tab:comparison-a}
        \resizebox{\textwidth}{!}{%
        \begin{tabular}{lcccc}
            \toprule
            Method & \shortstack{Classification\\(Top-1)} & \shortstack{Rotation\\($R^2$)} & \shortstack{Translation\\($R^2$)} & \shortstack{Colour\\($R^2$)} \\
            \midrule
            \multicolumn{5}{l}{\textcolor{gray}{\textit{Supervised Methods}}} \\
            ResNet-18 & 80.13 & 0.73 & 0.99 & 0.99 \\
            \midrule
            \multicolumn{5}{l}{\textcolor{gray}{\textit{Invariant \& Parameter Prediction Methods}}} \\
            VICReg & 74.71 & 0.39 & 0.81 & 0.50 \\
            SimCLR & 80.08 & 0.44 & 0.76 & 0.50 \\
           SimCLR + AugSelf & \textbf{80.86} & \textbf{0.69} & \textbf{0.85} & 0.51 \\
            \midrule
            \multicolumn{5}{l}{\textcolor{gray}{\textit{Equivariant Methods}}} \\
            SEN & 80.20 & 0.45 & 0.79 & 0.47 \\
           EquiMod& \textbf{80.76} & 0.43 & 0.72 & 0.38 \\
            SIE & 75.56 & 0.45 & 0.69 & 0.46 \\
            CapsIE & 75.20 & 0.60 & 0.85 & 0.03 \\
            \rowcolor{orange!15}EquiCaps & 76.91 & \textbf{0.73} & \textbf{0.88} & 0.05 \\
            \bottomrule
        \end{tabular}}
    \end{subtable}%
    \hfill
    \begin{subtable}[t]{0.49\textwidth}
        \centering
        \caption{
        Rotation \& Translation Pre-training. Equivariant methods are trained for both rotation and translation equivariance. We indicate the change relative to~\cref{tab:comparison-a}.
        }
        \label{tab:comparison-b}
        \resizebox{\textwidth}{!}{%
        \begin{tabular}{lcccc}
            \toprule
            Method & \shortstack{Classification\\(Top-1)} & \shortstack{Rotation\\($R^2$)} & \shortstack{Translation\\($R^2$)} & \shortstack{Colour\\($R^2$)} \\
            \midrule
            {\textcolor{gray}{\textit{Parameter Prediction Methods}}} \\
            SimCLR + AugSelf & 80.49 ($\downarrow$ 0.37) & 0.68 ($\downarrow$ 0.01) & 0.98 ($\uparrow$ 0.13) & 0.48 ($\downarrow$ 0.03) \\
            \midrule
            {\textcolor{gray}{\textit{Equivariant Methods}}} \\
            SEN & 80.24 ($\uparrow$ 0.04) & 0.45 (= 0.00) & 0.78 ($\downarrow$ 0.01) & 0.48 ($\uparrow$ 0.01) \\
            EquiMod &\textbf{80.55} ($\downarrow$ 0.21) & 0.45 ($\uparrow$ 0.02) & \textbf{0.98} ($\uparrow$ 0.26) & 0.37 ($\downarrow$ 0.01) \\
            SIE & 73.40 ($\downarrow$ 2.16) & 0.42 ($\downarrow$ 0.03) & \textbf{0.98} ($\uparrow$ 0.29) & 0.39 ($\downarrow$ 0.07) \\
            CapsIE & 72.88 ($\downarrow$ 2.32) & 0.33 ($\downarrow$ 0.27) & 0.96 ($\uparrow$ 0.11) & 0.05 ($\uparrow$ 0.02) \\
            \rowcolor{orange!15}EquiCaps & 77.88 ($\uparrow$ 0.97) & \textbf{0.71} ($\downarrow$ 0.02) & 0.91 ($\uparrow$ 0.03) & 0.02 ($\downarrow$ 0.03) \\
            \bottomrule
        \end{tabular}}
    \end{subtable}
\end{table*}

\section{Base-frame Translation Evaluation}\label{sec:baseframe}
\subsection{Rotation-Only Pre-Training}
\cref{tab:comparison-a} summarises the downstream performance of the learned representations on 3DIEBench-T when all equivariant methods are optimised only for rotation equivariance during pretraining. For clarity,~\cref{tab:comparison-a} uses the same pretrained weights and all the same metrics as~\cref{tab:allresults}; the only change is that the translation evaluation is recomputed in the base frame instead of the object frame.
Consistent with~\cref{tab:allresults}, EquiCaps again attains the highest translation performance when we perform base-frame evaluation.
Thus, even without direct translation supervision, EquiCaps achieves the best translation equivariance in both object and base frames.

\subsection{Rotation-Translation Pre-Training}

For completeness,~\cref{tab:comparison-b} reports results when both rotation and translation parameters are optimised. Similarly to the object frame setting, almost all methods exhibit an improvement in translation performance. 
Notably, EquiCaps is the only equivariant method that achieves rotation prediction within 0.02 $R^2$ of the supervised upper bound. 
In contrast, although the other equivariant methods improve their base frame translation prediction performance, their rotation performance remains well below EquiCaps and is similar to the invariant baselines. This suggests that their predictors offer no meaningful gain in rotational equivariance.
Overall, across all possible pre-training settings and evaluations, EquiCaps is the only method that consistently achieves high equivariance performance on both rotation and translation, demonstrating its generalisation ability under multi-geometric equivariance learning. 

\section{3DIEBench-T Dataset Generation}\label{sec:dataset-generation}

\begin{table}[t]
  \centering
  \caption{Parameter ranges for uniformly random object rotation, translation, and lighting in 3DIEBench-T. Tait–Bryan angles are used to define extrinsic object rotations, and the light’s position is specified using spherical coordinates.}
  \begin{tabular}{lcc}
    \hline
    Parameter        & Min. Value & Max. Value \\
    \hline
    Object rotation X       & $-\pi/2$           & $\pi/2$            \\
    Object rotation Y       & $-\pi/2$           & $\pi/2$            \\
    Object rotation Z       & $-\pi/2$           & $\pi/2$            \\
    Object translation X        & $-0.5$             & $0.5$              \\
    Object translation Y        & $-0.5$             & $0.5$              \\
    Object translation Z        & $-0.5$             & $0.5$              \\
    Floor   hue         & $0$                & $1$ \\
    Light hue & $0$ & $1$                \\
    Light $\theta$             & $0$                & $\pi/4$            \\
    Light $\phi$               & $0$                & $2\pi$             \\
    \hline
  \end{tabular}
  \label{tab:parameter_ranges}
\end{table}
\subsection{Generation Protocol}

3DIEBench-T extends the original settings of the 3DIEBench dataset by incorporating object translations. The dataset consists of 52,472 object instances across 55 classes from ShapeNetCoreV2 which are originally sourced from 3D Warehouse.
For each instance, 50 random views are rendered, sampled from a uniform distribution over the parameter ranges listed in~\cref{tab:parameter_ranges}, producing images at a resolution of 256$\times$256 pixels. For each view, the object is first translated by $t$ and then rotated by $R$, thus, the object's final base translation is $Rt$. For completeness, we evaluate the translation prediction task in both frames by using $t$ for the object frame and the resulting $Rt$ for the base frame.

We follow the original settings of 3DIEBench in which the rotation ranges are constrained to make the task more controllable, and the lighting angle is adjusted to ensure that shadows do not provide a trivial shortcut for the model.
Translation ranges are restricted so that objects do not move outside the camera’s view. 
We also store the transformation parameters as latent information alongside each image, facilitating equivariant tasks.  
Following the 3DIEBench split protocol, 80\% of the objects are used for training, while the remaining 20\%, unseen during training but sampled from the same transformation distribution, form the test set.
Generating the entire dataset requires approximately 44 hours on 12 NVIDIA A100 80GB GPUs, though the process can run in a single script.
\clearpage
\newpage
 
\subsection{Supplementary Image Samples}
\vspace{1.7cm}
\begin{figure}[H]
  \centering
  \begin{minipage}{\textwidth}
    \centering
    \includegraphics[width=\textwidth]{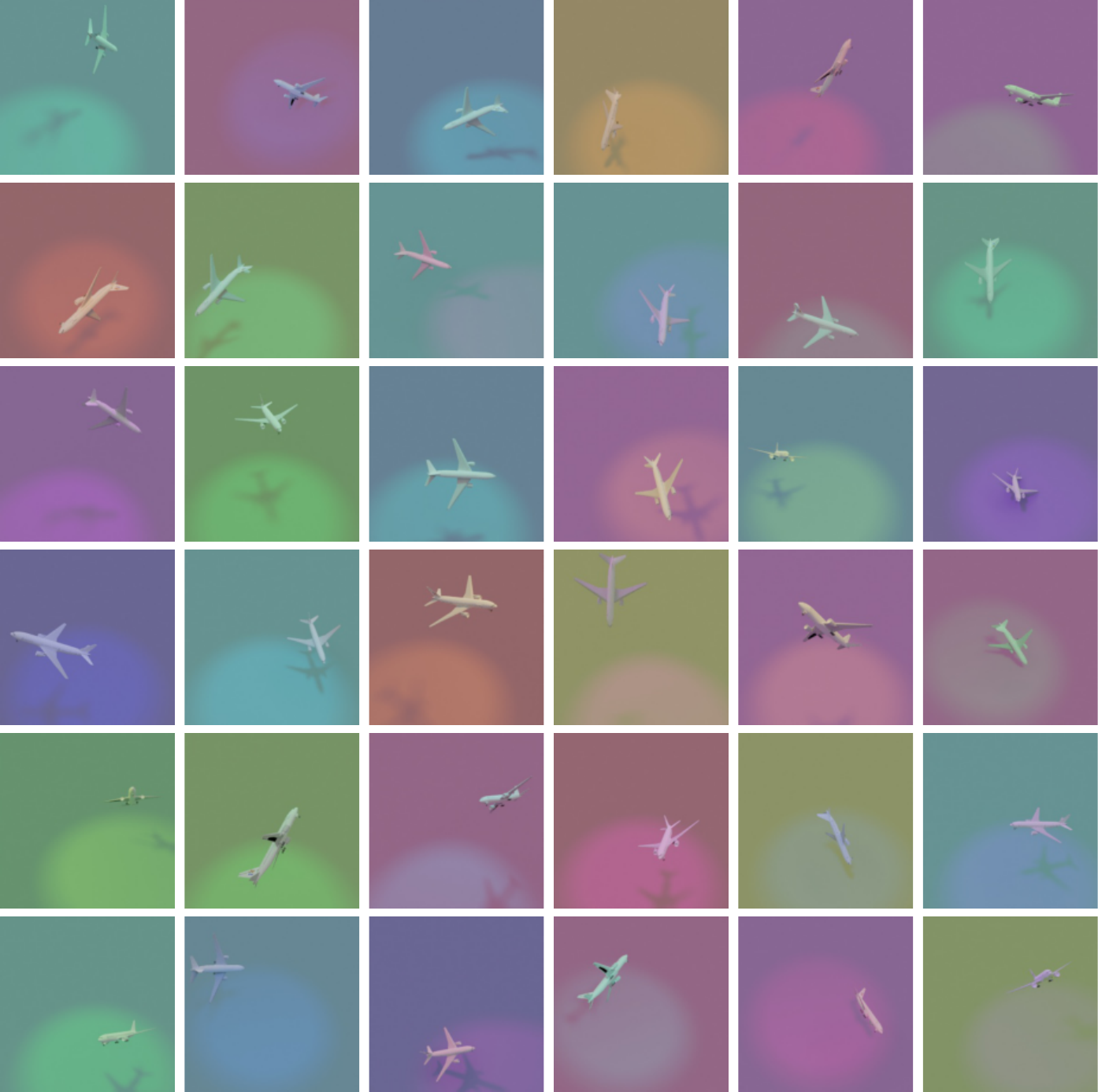}
    \captionof{figure}{Samples of an object instance from the 3DIEBench-T dataset.}
    \label{fig:plane-samples}
  \end{minipage}
\end{figure}

\clearpage
\begin{figure*}[]
    \centering
    \includegraphics[width=\linewidth]{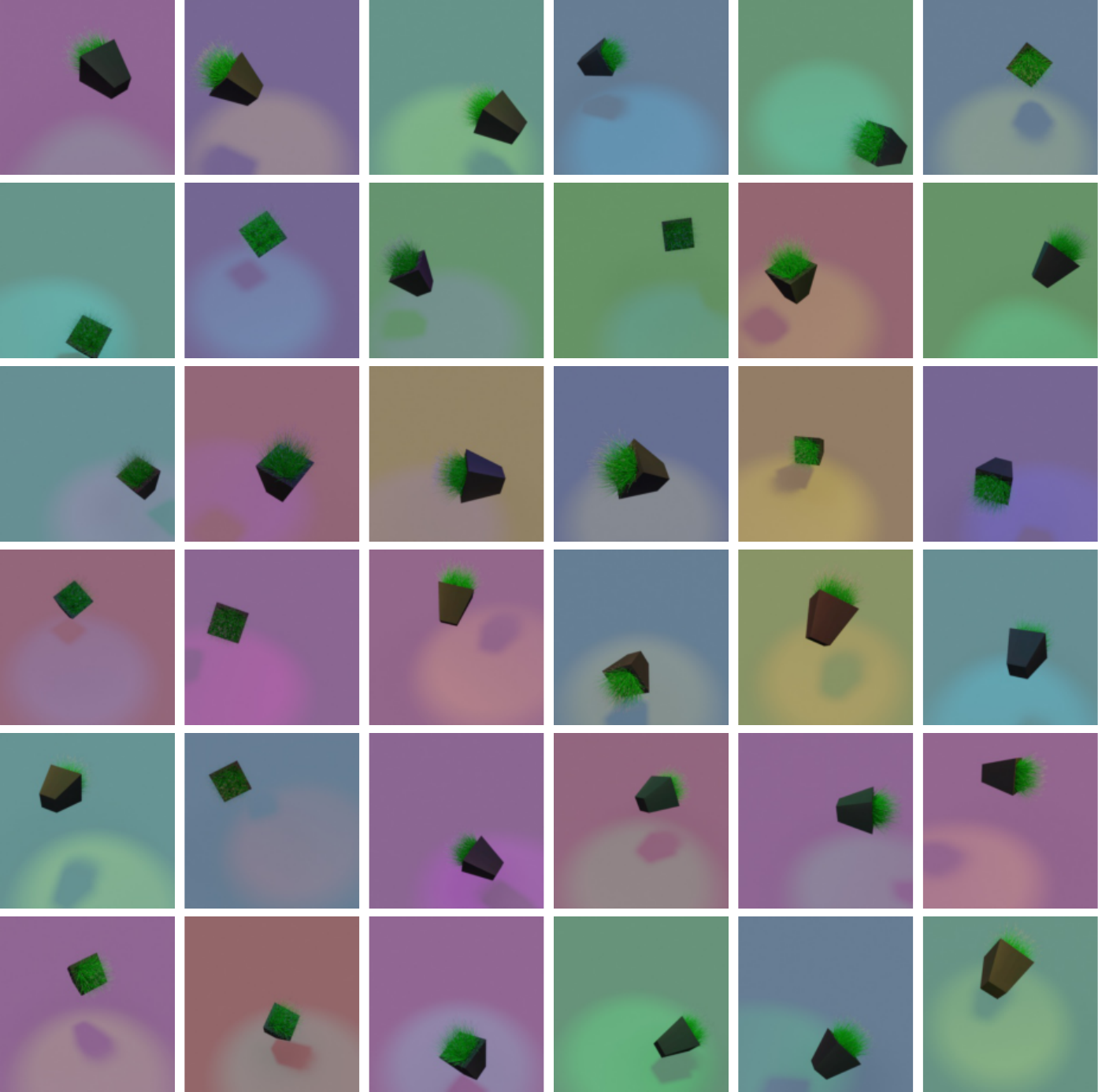}
    \caption{Samples of an object instance from the 3DIEBench-T dataset.}
    \label{fig:enter-label}
\end{figure*}
\clearpage
\begin{figure*}[]
    \centering
    \includegraphics[width=\linewidth]{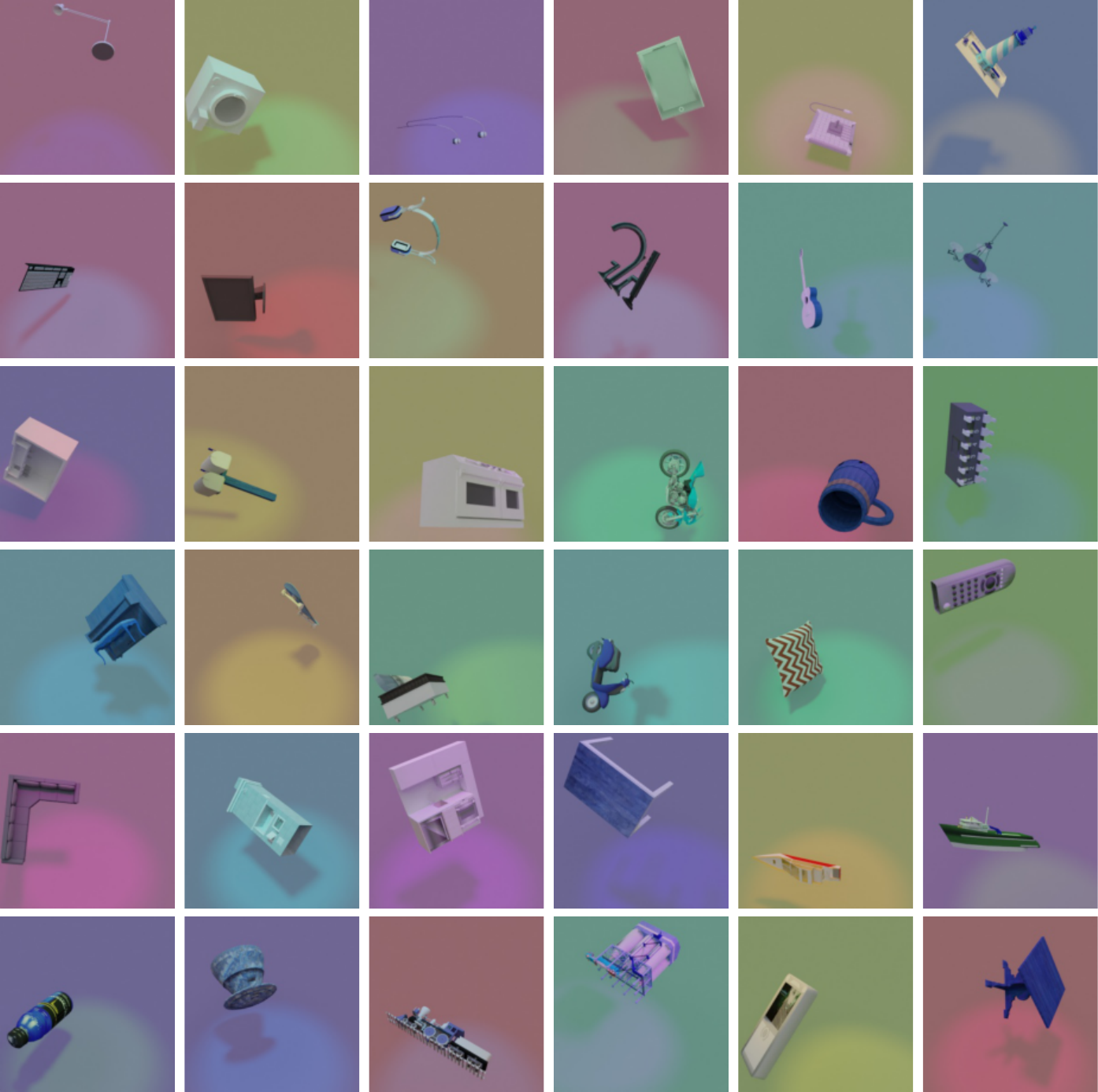}
    \caption{Samples of object instances from the 3DIEBench-T dataset.}
    \label{fig:enter-label}
\end{figure*}

\end{document}